\documentclass[10pt,twocolumn,twoside] {IEEEtran}

\usepackage[section]{algorithm}
\usepackage{algorithmic}
\usepackage{array}
\usepackage{amsmath}
\usepackage{amssymb}
\usepackage{amsfonts}
\usepackage{amsthm}
\usepackage{booktabs}
\usepackage{color}
\usepackage{graphicx}
\usepackage{subfigure}
\usepackage{multirow}
\usepackage{threeparttable}
\usepackage{hyperref}
\usepackage{makecell}
\usepackage{fmtcount}

\newcolumntype{L}[1]{>{\raggedright\let\newline\\\arraybackslash\hspace{0pt}}m{#1}}
\newcolumntype{C}[1]{>{\centering\let\newline\\\arraybackslash\hspace{0pt}}m{#1}}
\newcolumntype{R}[1]{>{\raggedleft\let\newline\\\arraybackslash\hspace{0pt}}m{#1}}

\numberwithin{equation}{section}

\theoremstyle{remark}

\newcommand{\cA}{\mathcal{A}}

\newcommand{\imgX}{u}
\newcommand{\imgY}{f}

\newcommand{\patchx}{x}
\newcommand{\patchy}{y}

\newcommand{\eg}{e.g.}

\newcommand{\cI}{\mathcal{I}}

\newcommand{\cL}{\mathcal{L}}
\newcommand{\cD}{\mathcal{D}}

\newcommand{\vt}{\vartheta}

\newcommand{\st}{\mathrm{subject~ to}\;}

\newcommand{\R}{\mathbb{R}}

\newcommand{\norm}[2][]{\|{#2}\|_{#1}}

\newcommand{\diag}{\mathrm{diag}}

\newcommand{\eps}{\varepsilon}

\newcommand{\scal}[2]{\left\langle #1,#2 \right\rangle}
\newcommand{\la}{\langle}
\newcommand{\ra}{\rangle}

\newcommand{\suml}[2]{\sum\nolimits_{#1}^{#2}}

\newcommand{\tabincell}[2]{\begin{tabular}{@{}#1@{}}#2\end{tabular}}

\title{Insights into analysis operator learning: From patch-based
  sparse models to higher-order MRFs} \author{Yunjin~Chen,
  Ren{\'e}~Ranftl and Thomas~Pock \thanks{Y.J. Chen is with the
    Institute for Computer Graphics and Vision, Graz University of
    Technology, Inffeldgasse 16, A-8010 Graz, Austria, as well as the
    College of Opt-Electronic Science and Engineering, National
    University of Defense Technology, Changsha 410073, Hunan, China.}
  \thanks{R. Ranftl and T. Pock are with the Institute for Computer
    Graphics and Vision, Graz University of Technology, Inffeldgasse
    16, A-8010 Graz, Austria.  e-mail: (\{cheny, ranftl,
    pock\}@icg.tugraz.at).}  \thanks{This work was supported by the
    Austrian Science Fund (FWF) under the China Scholarship Council
    (CSC) Scholarship Program and the START project BIVISION,
    No. Y729.}}

\begin{document}

\markboth{IEEE Transaction on Image Processing,~Vol.~xx, No.~xx, 20xx}
{Chen \MakeLowercase{\textit{et al.}}: Insights into analysis operator learning}
\maketitle

\begin{abstract}
  This paper addresses a new learning algorithm for the recently
  introduced co-sparse analysis model.  First, we give new insights
  into the co-sparse analysis model by establishing connections to
  filter-based MRF models, such as the Field of Experts (FoE) model of
  Roth and Black. For training, we introduce a technique called
  bi-level optimization to learn the analysis operators. Compared to
  existing analysis operator learning approaches, our training
  procedure has the advantage that it is unconstrained with respect to
  the analysis operator.  We investigate the effect of different
  aspects of the co-sparse analysis model and show that the sparsity
  promoting function (also called penalty function) is the most
  important factor in the model.  In order to demonstrate the effectiveness of our
  training approach, we apply our trained models to various classical
  image restoration problems. Numerical experiments show that our trained models clearly
  outperform existing analysis operator learning approaches and are on
  par with state-of-the-art image denoising algorithms.  Our approach
  develops a framework that is intuitive to understand and easy to
  implement.
\end{abstract}


\IEEEpeerreviewmaketitle

\section{Introduction}
\IEEEPARstart{O}{}ne of the most successful approaches to solve
inverse problems in image processing is to minimize a suitable energy
functional whose minimizer provides a trade-off between a smoothness
term and a data term. In a Bayesian framework, energy minimization can
also be interpreted as finding the Maximum-a-Posteriori (MAP)
estimate. Hence, the smoothness term is related to the image prior
distribution and the data term is related to the data likelihood.
Some classical smoothness terms are the squared $\ell_2$ norm of the
image gradients, the $\ell_1$ norm of the image gradients (total
variation), wavelet sparsity or more general MRF priors. Among the
overwhelming number of priors in the literature, sparse
representations have attracted remarkable attention in the last
decade.  Sparse representation models have been intensively
investigated and widely used in various image processing applications
such as image denoising, image inpainting, super resolution, etc. See
for example \cite{KSVDdenoising2006, LSSC, SCinpainting} and
references therein.
\subsection{Patch-based synthesis and analysis models}
Historically, sparse representations refer to the so-called sparse
synthesis model. In the synthesis-based models, a signal $\patchx \in
\R^m$ (a patch) is called sparse over a given dictionary $D \in
\R^{m\times n}$ with $m\le n$, when it can be composed of a linear
combination of only a few atoms from dictionary $D$.  This is
formulated as the following minimization problem:
\begin{equation}\label{synthesismodel}
x^* = D\alpha^*; \alpha^* = \arg\min\limits_{\alpha \in \R^n} 
\phi(\alpha)  + 
\frac{\lambda}{2}\norm[2]{D\alpha - f}^2 \,, 
\end{equation}
where $f \in \R^m$ is the observed patch, $\alpha \in \R^n$ is the
coefficient vector and $\phi$ is the penalty function.  In order to
induce sparsity in the representation, typical penalty functions
include $\ell_p$ norms with $p \in\{0,1\}$ or logarithmic functions
such as $\text{log}(1+|z|^p)$, with $p \in\{1,2\}$. The synthesis
model has been intensively studied in the past decade, including
global and specialized dictionary learning algorithms and applications
to various image processing tasks, see
\cite{KSVDdenoising2006,KSVD2006,Lee2006,MairalBPS09,MairalES08} for
examples.

However, there is another viewpoint to consider sparse
representations, which is the so-called co-sparse analysis
model~\cite{EladAnalysisVSSynthesis}.  The objective of a co-sparse
model is to pursue a linear operator $A \in \R^{n\times m}$, such that
the resulting coefficient vector $A\patchx \in \R^n$ is expected to be
sparse.  In the framework of MAP inference, the co-sparse analysis
model is given as the following minimization problem:
\begin{equation}\label{analysispatch}
\patchx^* = \arg\min\limits_{\patchx \in \R^m} 
\phi(A\patchx)  + 
\frac{\lambda}{2}\norm[2]{\patchx - f}^2 \,, 
\end{equation}
where $A$ is the so-called analysis operator, $\phi$ is again a
sparsity promoting function as mentioned above and $f \in \R^m$ is the
observed patch.  Note that both the analysis model and the synthesis
model become equivalent if $D$ is invertible. However, the analysis
model is much less investigated compared to the well-known synthesis
model, but it has been gaining more and more attention in recent
years~\cite{HaweAnalysisLearning,FadiliAnalysisLearning,
  RubinsteinPE13}.

\subsection{Patch-based analysis operator learning}
In the case of the synthesis model, the learning of an optimized
dictionary has become ubiquitous. However, in analysis-based models,
fixed operators inspired from variational methods such as the discrete
total variation have been used for a long time. It is only recently
that people started to develop customized algorithms to learn in some
sense optimal analysis operators.

Existing algorithms mainly concentrate on the patch-based training
strategy. Given a set of $S$ training samples $ {Y = [\patchy_1, \dots,
\patchy_S] \in \R^{m\times S}}$, where depending of the training
procedure, each sample is a noisy version or a clean version of an
image patch.  For the noisy version, $\patchy_i = \patchx_i + n_i$,
where $n_i$ is an additive zero-mean white Gaussian noise vector and
$\patchx_i$ is the clean signal.

For the noise-free training, the goal of the analysis operator learning is
to find a linear operator $A \in \R^{n\times m}$ with $m\le n$, such
that the coefficient vector $A\patchy_i$ is as sparse as possible.
This strategy can be formally expressed as the following optimization
problem.
\begin{equation}\label{noise free model}
  \ A^\star = \arg\min\limits_{A} \phi(AY). 
\end{equation}

For the noise aware case, the objective is to learn an optimal
analysis operator $A$, which enforces the coefficient vector
$A\patchx_i$ to be sparse, while $\norm[2]{\patchx_i - \patchy_i}^2
\leq \eps$ for each training sample ($\eps$ is an error tolerance, which is
derived from the noise level).  This requires solving a problem of the
form
\begin{align}\label{noise aware model}
\{A^\star, X\} = 
\arg\min\limits_{A, X} \phi(AX), \nonumber\\ \st 
\norm[2]{\patchx_i - \patchy_i}^2 \leq \eps. 
\end{align}
Using a Lagrange multiplier $\lambda > 0$ this can be equivalently
expressed as 
\begin{equation}\label{noise aware model2}
\ \{A^\star, X\} = 
\arg\min\limits_{A, X} \phi(AX) +  
\frac{\lambda}{2}\norm[F]{X - Y}^2, 
\end{equation}
where $\phi$ is again a sparsity promoting function and
$\norm[F]{\cdot}$ denotes the Frobenius norm.

Unfortunately, the above optimization problems suffer from the problem
of trivial solutions.  Indeed, if no constraints are imposed on $A$,
it is easy to see that the trivial solution $A \equiv 0$ is the global
minimizer of \eqref{noise free model}, \eqref{noise aware model} and
\eqref{noise aware model2}.  A possible solution to exclude the
trivial solution is to impose additional assumptions on $A$, i.e.,
restricting the solution set to an admissible set $\mathcal{C}$. The
following constraints have been investigated in
\cite{YahoobiAnalysisLearning} and \cite{HaweAnalysisLearning}:
\begin{itemize}
\item[(i)] row norm constraints. All the rows of $A$ have the same
  norm, i.e., $\norm[2]{A_i} = c$ for the $i^{th}$ row of operator
  $A$.
\item[(ii)] row norm + full rank constraints. The analysis operator
  $A$ has full rank, i.e., $rk(A) = m$.
\item[(iii)] tight frame constraints. The admissible set of this
  constraint is the set of tight frame in $\R^{n\times m}$, i.e.,
  $A^TA = \cI_m$, where $\cI_m$ is the identity operator in $\R^m$.
\end{itemize}

As pointed out in \cite{YahoobiAnalysisLearning}, each individual
constraint presented above does not lead to satisfactoy results.
Therefore, in \cite{YahoobiAnalysisLearning,YahoobiNoiseAware} a
constraint called the Uniform Normalized Tight Frame (UNTF) was
proposed, which is a combination of the unit row norm and the tight
frame constraint. The authors of \cite{HaweAnalysisLearning} employed
a constraint combining the unit row norm and the full rank constraint
with an additional consideration that the analysis operator $A$
doesn't have trivially linear dependent rows, i.e., $A_i \ne A_j$ for
$i \ne j$.

In \cite{YahoobiAnalysisLearning}, Yaghoobi et al. employed the convex
$\ell_1$-norm, i.e., $\phi(AY) = \norm[1]{AY}$, as sparsity promoting
function and the UNTF constraint to solve problem \eqref{noise free
  model}.  In \cite{YahoobiNoiseAware}, the same authors proposed an
extension of their previous algorithm that simultaneously learns the
analysis operator and denoises the training samples. The improved
algorithm solves the problem~\eqref{noise aware model2} by alternating
between updating the analysis operator and denoising the training
samples.  They gave some preliminary image denoising results by
applying the learned operator to natural face images.

In \cite{HaweAnalysisLearning}, Hawe et al. exploited the above
constraints - full rank matrices with normalized rows, and a
non-convex sparsity measurement function called the mixed
$(p,q)$-pseudo-norm to minimize problem \eqref{noise free model}.
They employed a conjugate gradient method on manifolds to solve this
optimization problem.  Their experimental results for classical image
restoration problems show competitive performance compared to
state-of-the-art techniques.

Rubinstein et al. \cite{RubinsteinKSVD} presented an adaption of the
widely known K-SVD dictionary learning method \cite{KSVDdenoising2006}
to solve the problem \eqref{noise aware model} directly based on the
$\ell_0$ quasi-norm, i.e., $\phi(AY) = \norm[0]{AY}$. Unfortunately,
there are only synthetic experiments and examples based on piece-wise
constant images considered in their work. The same authors presented
some preliminary natural image denoising results in their later
work~\cite{RubinsteinPE13}.  However, it turns out that the
performance of the learned analysis operator is inferior to the
synthesis model~\cite{KSVDdenoising2006}.

Ophir et al. \cite{OphirSequentialLearning} proposed a simple analysis
operator learning algorithm, where analysis \textquotedblleft
atoms\textquotedblright ~are learned sequentially by identifying
directions that are orthogonal to a subset of the training data.

Apart from the above analysis operator learning algorithms, Peyr{\'e}
and Fadili proposed an attractive learning approach in~\cite{FadiliAnalysisLearning}.  
They considered the analysis operator
from a particular viewpoint. They interpreted the behavior of the
analysis operator as a convolution with some finite impulse response
filters. Keeping this idea in mind, they formulated the analysis
operator learning as a bi-level programming problem
\cite{bileveloverview} which was solved using a gradient descent
algorithm. However, their work only considered a simple case - one
filter and 1D signals.  Following this direction, a preliminary
attempt to apply this idea to 2D image processing was done in
\cite{Chennips2012}.
\subsection{Motivation and contributions}
Among the existing algorithms for analysis operator learning, only few
prior works have been evaluated based on natural images
\cite{YahoobiNoiseAware,RubinsteinPE13,HaweAnalysisLearning}.
Moreover, most of these algorithms have to impose a non-convex
constraint on the analysis operator $A$, making the corresponding
optimization problems hard to solve.  Thus a question arises: Is it
possible to introduce a more principled technique to learn optimized
analysis operators without the need to impose additional constraints
on the operators?

In this paper, we give an answer to this question.  First, we extend
the patch-based analysis model to a global image regularization term,
which allows to consider also more general inverse problems such as
image deconvolution and image inpainting. Then, we show that this
model is equivalent to higher-order filter-based MRF models such as
the FoE model \cite{RothFOE2009}.  Motivated by this observation, we
apply a loss-function based training scheme~\cite{SamuelFoE} and show
that this approach excludes the trivial solution of the analysis
operator learning problem without imposing any additional constraints.
Furthermore, we carefully investigate the effect of different aspects
of the analysis based model.  We show that the choice of the sparsity
promoting function is the most important aspect.  We present various
experimental results for standard image restoration problems to
demonstrate the effectiveness of our training model. Numerical results
show that our trained model significantly outperforms existing
analysis operator based models and is on par with specialized image
denoising algorithms while being computationally very
efficient. Therefore, our training procedure provides an attractive
alternative to existing approaches.

A shorter version of this paper was presented in GCPR \cite{chen}.

\subsection{Notation}
In this paper, our model presents a global prior over the entire image
instead of small image patches.  In order to distinguish between a
small patch and an entire image, we represent a square patch (patch
size: $\sqrt{m} \times \sqrt{m}$) by $\patchx \in \R^m$, and an image
(image size: $M \times N$, with $m \ll M, m \ll N$) by $\imgX \in
\R^{MN}$. We denote the patch-based synthesis dictionary and analysis
operator by $D \in \R^{m\times n}$ and $A \in \R^{n\times m}$ with
$m\le n$, respectively.  Furthermore, when the analysis operator $A$ is
applied to the entire image $\imgX$, we use the common sliding-window
fashion to compute the coefficients $Ax$ for all $MN$ patches in the
image.  This result is equivalent to a multiplication of a highly
sparse matrix $\cA \in \R^{(n \times MN) \times MN}$ with the image
$\imgX$, i.e., $\cA \imgX$.  We can group $\cA$ to $n$ separable
sparse matrices $\{\cA_1,\dots,\cA_n\}$, where $\cA_i \in \R^{MN
  \times MN}$ is associated with the $i^{th}$ row of $A$ ($A_i$). If
we consider $A_i$ as a 2-D filter ($\sqrt{m} \times \sqrt{m}$), we
have: $\cA_i u$ is equivalent to the result of convolving image $u$
with filter $A_i$.

\section{Insights into analysis based models}\label{insights}
In this section, we first show the equivalence between the patch-based
analysis model and filter-based probabilistic image patch modeling -
Product of Experts (PoE)~\cite{WellingHO02,POE}. Then we extend the
patch-based analysis model to the image-based model and show
connections to higher order MRFs~\cite{RothFOE2009}.

\subsection{Equivalence between the patch-based analysis model and the
  PoE model}
The patch-based analysis model in \eqref{analysispatch} focuses on
modeling small image patches, which is formulated as a matrix-vector
multiplication ($Ax$). This procedure can be interpreted as projecting
a signal $x$ (an image patch) onto a set of linear components
$\{A_i\}_{i=1}^{n}$, where each component $A_i$ is a row of the matrix
$A$. Note that projecting an image patch onto a linear component
($A_ix$) is equivalent to filtering the patch with a linear filter
given by $A_i$.

The PoE model provides a prior distribution on small image patches by
taking the product of several expert distributions, where each expert
works on a linear filter and the expert function. The PoE model is
formally written as $p(x) = \frac{1}{Z(\Theta)} \text{exp}(-E_{PoE}
(x, \Theta))$ with
\begin{equation}\label{poe}
E_{\text{PoE}}(x,\Theta) = - \suml{i=1}{n} \text{log} \rho_i (A_i x), 
\end{equation}
where $\rho_i$ is the potential function, $Z(\Theta)$ is the
normalization and $\Theta$ are the parameters of this model.

Comparing the analysis prior given in \eqref{analysispatch}, ${\phi(Ax)
= \suml{i=1}{n}\phi_i(A_ix)}$ with the above PoE model, we can see they
are actually the same model if we choose the penalty function as
$\phi_i = -\text{log}\rho_i$.  In this case, if we
consider the analysis operator learning problem based on the strategy
which focuses on the modeling of small image patches rather than
defining a prior model over an entire image, the learning problem is
tantamount to learning filters in the PoE model.

\subsection{From patch-based to image-based model}
Patch-based models are only valid for the reconstruction of a single
patch.  When they are applied to full image recovery, a common
strategy is patch averaging \cite{KSVDdenoising2006}.  All the patches
in the entire image are treated independently, reconstructed
individually and then integrated to form the final reconstruction
result by averaging the overlapping regions.  While this method is
simple and intuitive, it clearly
ignores the coherence between over-lapping patches, and thus misses
global support during image reconstruction. To overcome these
drawbacks, an extension to the whole image is necessary where patches
are not treated independently but each of them is a part of the image.

A promising direction to formulate an image-based model is to make use
of the formalism of higher-order MRFs which enforce coherence across
patches \cite{HaweAnalysisLearning}.  The basic idea is to modify the
patch-based analysis model in \eqref{analysispatch} such that all 
possible patches in the entire image and the corresponding 
coefficient vectors $Ax$ are considered at once. This leads to an
image-based prior model of the form:
\begin{equation}\label{energyfunction}
E_{\text{prior}}(\imgX) = \suml{p=1}{N_p} \phi(AP_p \imgX), 
\end{equation}
where $u$ is an image of size $M \times N$, $N_p = N \times M$,
$\phi(AP_p \imgX) = \suml{i=1}{n} \phi((AP_p \imgX)_i)$ and $P_p \in
\R^{m \times N_p}$ is a sampling matrix extracting the patch at pixel
$p$ in image $u$.  For the patches at the image boundaries, we extract
patches by using symmetric boundary conditions.

A key characteristic of the model~\eqref{energyfunction} is that it
explicitly models the overlapping of image patches, which are highly
correlated.  Intuitively, it is a better strategy for image modeling
compared to the patch averaging approach. In
Subsection~\ref{denoising} we will provide experimental results
to support this claim.
\subsection{Equivalence between the image-based analysis model and the
  FoE model}
If we consider in \eqref{energyfunction} each row of $A$ ($A_i$) as a
2-D filter ($\sqrt{m} \times \sqrt{m}$), we can rewrite this term as
\begin{equation}\label{prior}
E_{\text{prior}}(\imgX) = \suml{p=1}{N_p} \suml{i=1}{n} \phi(A_i * \imgX_p), 
\end{equation}
where $(A_i * \imgX_p)$ denotes the result of convolving the patch at
pixel $p$ with filter $A_i$.  After having a closer look at this prior
term, interestingly we find that it is the same as the FoE model
proposed by Roth and Black \cite{RothFOE2009}.  The FoE models the
prior probability of an image by using a set of linear filters and a
potential function.  The probability density function for the entire
image is written as $p (u) = \frac{1}{Z(\Theta)} \text{exp}(-E_{FoE}
(u, \Theta))$ with
\begin{equation}\label{foe}
E_{\text{FoE}}(\imgX,\Theta) = -\suml{p=1}{N_p} \suml{i=1}{n} \text{log} \rho_i (A_i * \imgX_p), 
\end{equation}
where $\rho_i$ is the potential function, $Z(\Theta)$ is the
normalization and $\Theta$ is a vector holding the parameters of this
model.  Based on the observation that responses of linear filters
applied to natural images typically exhibit heavy tailed distribution,
two types of heavy tailed potential functions, the Student-t
distribution (ST) and generalized Laplace distribution (GLP), are
commonly considered:
\begin{equation}\label{potentialfunctions}
\begin{cases}
\rho_i(z;p_i) = (1 + z^2)^{-p_i}\hspace{0.45cm} (\text{ST})\\
\rho_i(z;p_i) = e^{-|z|^{p_i}} \hspace{1.2cm}(\text{GLP}).
\end{cases}
\end{equation}
Comparing \eqref{prior} and \eqref{foe}, we can see that they are
exactly the same if we choose the penalty function $\phi_i =
-\text{log} \rho_i$,
\begin{equation}\label{equivalence}
\begin{cases}
\rho_i(z;p_i) = (1 + z^2)^{-p_i} \Longleftrightarrow \phi_i = p_i \text{log}(1+z^2)\\
\rho_i(z;p_i) = e^{-|z|^{p_i}} \hspace{0.75cm}\Longleftrightarrow \phi_i = |z|^{p_i}\,.
\end{cases}
\end{equation}
Note that these choices lead to commonly used non-convex sparsity
promoting functions \cite{WellingHO02,HaweAnalysisLearning}.

In conclusion, the FoE model can be seen as an extension of the
co-sparse analysis model from a patch-based formulation to an
image-based formulation. It comes along with the advantage of
inherently capturing the coherence between overlapping patches which
has to be enforced explicitly in patch-based models. As we will see
in the next section, the image-based model also allows to learn
optimized analysis operators without the need for additional
constraints.

\section{Learning}\label{trainingmodel}
In this section, we first present a loss-based training procedure
based on bi-level optimization.  Our algorithm is closely related to
the algorithm proposed in \cite{SamuelFoE} but we propose to solve the
lower level problems with high accuracy which leads to improved
gradient directions for minimizing the loss function with respect to
the model parameters.  As a result of this seemingly minor
modification, we achieve significantly better results compared to
previous work.  For more details about the refined training algorithm
we refer to \cite{chen}.

\subsection{Bi-level optimization}\label{model}

Existing approaches to learn the parameters in the FoE model fall into
two main types: (1) probabilistic learning using sampling-based
algorithms, e.g., \cite{RothFOE2009, GaoCVPR2010, GaoDAGM2012}; (2)
bi-level training based on MAP estimation, e.g.,
\cite{DomkeAISTATS2012, Barbu2009, SamuelFoE, TappenVariational}.
Reviewing all algorithms is beyond the scope of this paper. Here we
focus on the the bi-level training scheme and refer the interested
reader to \cite{chen} for a survey.

Bi-level optimization is a popular and effective technique for
selecting hyper parameters, cf.~\cite{DoFN07,TappenSDL08,
  pockbilevel}. The problem of learning the analysis operator can be
written as the following bi-level optimization problem
\begin{equation}\label{bilevel}
\begin{cases}
\arg\min\limits_{\vartheta}L(u^*(\vartheta),g)\\
\st u^*(\vartheta) = \arg\min\limits_{u}E(u,f,\vt)\,
\end{cases}
\end{equation}
Given a noisy observation $f$ and the ground truth $g$, our goal is to
find optimal hyper parameters $\vartheta$ such that the minimizer
$u^*(\vartheta)$ of the lower level problem $E(u,f,\vt)$ minimize the
higher level problem $L(u^*(\vartheta),g)$. In our case, the hyper
parameters will be used to parametrize the analysis operators as well
as the potential functions, the lower level problem is given by the
energy function and the higher level problem is given by a certain
loss function that compares the solution of the lower level problem
with the ground truth solution.

First, we consider the lower level problem. Treating the analysis
operator $A \in \R^{n \times m}$ in \eqref{prior} as $n$ filters with
dimension $\sqrt{m} \times \sqrt{m}$, the analysis model for image
denoising is given by
\begin{align}\label{analysismodel}
\imgX^* &= \arg\min\limits_{\imgX}
E(\imgX,f,A) \nonumber\\
&= \suml{i=1}{n} \alpha_i \phi(\cA_i \imgX) + \frac{\lambda}{2}\|\imgX-\imgY\|_2^2, 
\end{align}
where $\phi(\cA_i \imgX) = \suml{p=1}{N_p} \phi ((\cA_i \imgX)_p)$,
$\cA_i$ is an $N_p \times N_p$ highly sparse matrix, which makes the
convolution of the filter $A_i$ with a two-dimensional image $u$
equivalent to the product of the matrix $\cA_i$ with the
vectorization of $u$, i.e., $A_i * u \Leftrightarrow \cA_i u$,
and $\alpha_i \geq 0$ is the weight parameter associated to the filter~$\cA_i$. 
In our training model, we express the filter $\cA_i$ as a
linear combination of a set of basis filters $\{B_1,\cdots,B_{N_B}\}$,
i.e.,
\begin{equation}\label{linear}
\cA_i = \suml{j=1}{N_B}\beta_{ij}B_j. 
\end{equation}

The loss function is defined to penalize the difference (loss) between
the optimal solution of the energy minimization problem and the
ground-truth. In this paper, we make use of the following
differentiable function as in \cite{SamuelFoE}:
\begin{equation}\label{lossfunction}
L(u^*,g) = \frac{1}{2}\|u^* - g\|_2^2, 
\end{equation}
where $g$ is the ground-truth image and $u^*$ is the minimizer of
energy function \eqref{analysismodel}.  This loss function has an
interpretation of pursuing as high PSNR as possible.

Given the training samples $\{f_s,g_s\}_{s=1}^S$, where $g_s$ and
$f_s$ are the $s^{th}$ clean image and the associated noisy version
respectively, our aim is to learn an optimal analysis operator or a
set of filters which are defined by parameters $\vt = (\alpha, \beta)$
(we group the coefficients $\beta_{ij}$ and weights $\alpha_i$ into a
single vector~$\vt$), such that the overall loss function for all
samples is as small as possible. Therefore, our learning model is
formulated as the following bi-level optimization problem:
\begin{equation}\label{learningmodel}
\begin{cases}
\min\limits_{\alpha \geq 0, \beta}L(u^*(\alpha,\beta)) = 
\suml{s=1}{S}\frac{1}{2}\|u_s^*(\alpha,\beta) - g_s\|_2^2\\
\text{where}~
u_s^*(\alpha,\beta) = \arg\min\limits_{u}
\suml{i=1}{n} \alpha_i \phi(\cA_i \imgX) + \frac{1}{2}\|\imgX-\imgY_s\|_2^2. 
\end{cases}
\end{equation}
We eliminate $\lambda$ for simplicity since it can be incorporated
into the weights $\alpha$.  Our analysis operator training model has
two advantages over existing analysis operator learning algorithms.
\begin{itemize}
\item[(a)]\textit{It is completely unconstrained with respect to the
    analysis operator $A$.}  Normally, existing approaches such as
  \cite{YahoobiAnalysisLearning,YahoobiNoiseAware,HaweAnalysisLearning}
  have to impose some non-convex constraints over the analysis
  operator. On the one hand, this makes the corresponding optimization
  problem difficult to solve, and on the other hand it decreases the
  probability of learning a meaningful analysis operator, because as
  indicated in \cite{YahoobiAnalysisLearning}, there is no evidence to
  prove that the introduced constraints are the most suitable choices.
  The reason why constraints are indispensable for these approaches
  lies in the need to exclude the trivial solution $A = 0$.  However, looking back
  at our training model, this trivial solution can be avoided
  naturally.  If $A = 0$, the optimal solution of the lower-level
  problem in \eqref{learningmodel} is certainly $\imgX_s^* = \imgY_s$,
  which makes the loss function still large; thus this trivial
  solution is not acceptable in that the goal of our model is to
  minimize the loss function. Therefore, the optimal operator $A$ must
  comprise some meaningful filters such that the minimizer of the
  lower-level problem is close to the ground-truth.
\item[(b)]\textit{The learned analysis operator inherently captures the properties of overlapping patches.} 
In \cite{KSVDdenoising2006,RubinsteinKSVD,RubinsteinPE13,
YahoobiAnalysisLearning}, their approaches present a patch-based prior, 
and thus for global reconstruction of an entire image, the common strategy consists of two stages: 
(i) extract overlapping patches, reconstruct them individually by synthesis-prior or analysis-prior based model, 
and (ii) form the entire 
image by averaging the final reconstruction results in the overlapping regions. This strategy clearly misses 
global support during the 
reconstruction process; however our approach can overcome these drawbacks.
\end{itemize}

In the work of \cite{HaweAnalysisLearning}, the authors employ the
patch-based model to train the analysis operator, but use it in the
manner of an image based model. Clearly, if the final intent is to use
the analysis operator in an image-based model, a better strategy is to
train it also in the same framework.

\subsection{Solving the bi-level problem}\label{calgrad}
In this subsection, we consider the bi-level optimization problem from
a general point of view.  For convenience, we only consider the case
of a single training sample and we show how to extend the framework to
multiple training samples in the end.

According to the optimality condition, the solution of the lower-level
problem in~\eqref{learningmodel} is given by $u^*$, such that
$\nabla_uE(u^*) = 0$. Therefore, we can rewrite
problem~\eqref{learningmodel} as following constrained optimization
problem
\begin{equation}\label{constrainedmodel}
\begin{cases}
\min\limits_{\alpha \geq 0, \beta}L(u(\alpha, \beta)) = 
\frac{1}{2}\|u(\alpha, \beta) - g\|_2^2\\
\st
\nabla_uE(u) = \suml{i=1}{n}\alpha_i \cA_i^T\phi'(\cA_i u) + u- f = 0, 
\end{cases}
\end{equation}
where $\phi'(\cA_i u) = (\phi'((\cA_i u)_1),\cdots,\phi'((\cA_i u)_p))^T \in \R^{N_p}$. 
Now we can introduce Lagrange multipliers and study the Lagrange function
\begin{multline}\label{Lagrange}
\cL(u,\alpha,\beta,p,\mu) = 
\frac{1}{2}\|u - g\|_2^2 + \scal{-\alpha}{\mu} \\ 
+ \la \suml{i=1}{n}\alpha_i \cA_i^T\phi'(\cA_i u) + u- f,p\ra, 
\end{multline}
where $\mu \in \R^{n}$ and $p \in \R^{N_p}$ are the Lagrange multipliers associated to 
the inequality constraint $\alpha \geq 0$ and the equality constraint 
in~\eqref{constrainedmodel}, respectively. 
Here $\scal{\cdot}{\cdot}$ denotes the standard inner product. 
Taking into account the inequality constraint $\alpha \geq 0$, the first-order necessary condition 
for optimality is given by 
\begin{equation}\label{optimality}
G(u,\alpha,\beta,p,\mu) = 0,
\end{equation}
where
\begin{equation*}
G(u,\alpha,\beta,p,\mu)=
\begin{pmatrix}
    (\suml{i=1}{n}\alpha_i \cA_i^T\cD_i \cA_i+ \cI)p + u-g\\
    (\la \cA_i^T\phi'(\cA_ix),p \ra)_{n \times 1}-\mu\\
    (\la B_j^T\phi'(\cA_ix) + \cA_i^T\cD_i B_j u, p \ra)_{N_c \times 1}\\
     \suml{i=1}{n}\alpha_i \cA_i^T \phi'(\cA_ix) + u- f\\
    \mu - \max(0,\mu-c\alpha)
\end{pmatrix}.
\end{equation*}
Wherein $\cD_i(\cA_i u) = \diag (\phi''((\cA_i u)_1),\cdots,\phi''((\cA_i u)_p)) \in \R^{N_p \times N_p}$, 
$(\la \cdot ,p \ra)_{N_c \times 1} = (\la (\cdot)_1,p \ra,\cdots,\la (\cdot)_r,p \ra)^T$, 
and $N_c = n \times N_B$. 
Note that the last equation is derived from the optimality condition for the inequality 
constraint $\alpha \geq 0$, which is expressed 
as $\alpha \geq 0, \mu \geq 0, \la \alpha, \mu \ra = 0$. It is easy to check that these 
three conditions are equivalent to 
$\mu - \max(0,\mu-c\alpha) = 0$ with $c$ to be any positive scalar and max operates 
coordinate-wise. 

In principle, we can continue to calculate the second derivatives of \eqref{Lagrange}, i.e., 
the Jacobian matrix of G, with which we can then employ a Newton's method to 
solve the necessary optimality system~\eqref{optimality} as in \cite{pockbilevel}. 
However, for this problem, calculating the Jacobian of G is computationally expensive; 
thus in this paper we do not consider the second derivatives and only make use of 
the first derivatives. An efficient Newton's method is subject to the future work. 

In our training model, what we are interested in is the parameters $\vt=\{\alpha,\beta\}$. 
We can reduce unnecessary variables in \eqref{optimality}
by solving for $p$ and $u$ in~\eqref{optimality}, and substituting them into the second and the third 
equation. We arrive at the following gradients of the 
loss function with respect to the parameters $\vt$:
\begin{equation}\label{overallderivative}
\begin{cases}
\nabla_{\beta_{ij}} L = 
- (B_j^T\phi'(\cA_i u) + \cA_i^T\cD_i B_j u)^T(H_E(u))^{-1} \left(u - g\right)
\\
\nabla_{\alpha_{i}} L = 
- (\cA_i^T\phi'(\cA_i u))^T(H_E(u))^{-1}
\left(u - g\right)
\\
\text{where} ~\nabla_uE(u) = \suml{i=1}{n}\alpha_i \cA_i^T\phi'(\cA_iu) + u- f = 0. 
\end{cases}
\end{equation}
$H_E(u)$ denotes the Hessian matrix of $E(u)$, 
\begin{equation}\label{hessian}
H_E(u) = \suml{i=1}{n}\alpha_i \cA_i^T\cD_i \cA_i+ \cI.
\end{equation}

Note that in~\eqref{overallderivative} we also eliminated the Lagrange multiplier $\mu$ associated to 
the inequality constraint $\alpha \geq 0$ since we utilize a quasi-Newton's method for optimization, 
which can easily handle this type of box 
constraints; therefore we do not need to consider the inequality constraint in the derivatives. 
The derivatives in \eqref{overallderivative} are 
equivalent to the results presented in~\cite{SamuelFoE}, which used implicit differentiation for
the derivation. 

Considering the case of $S$ training samples, in fact it turns out 
that the derivatives of the overall loss function in \eqref{learningmodel} with respect 
to the parameters $\vt$ are just the sum of gradients given in \eqref{overallderivative} with respect to all the training samples. 
\subsection{Bi-level learning algorithm}
\begin{algorithm}\caption{Bi-level learning algorithm for analysis operator training}\label{algo3.1}
\begin{itemize}
\item[(i)] Given training samples $\{f_s,g_s\}_{s=1}^S$, 
initialization of parameters $\vt^0 = \{\alpha^0, \beta^0\}$, let $l = 0$,
\item[(ii)] For each training sample, solve for $u_s^*(\vt_l)$
\begin{align}\label{temp}
&\suml{i=1}{n} \alpha^l_i \cA_i^T\phi'(\cA_i u_s^*) + u_s^*- f_s = 0, \nonumber\\ 
&\textrm{where } \cA_i = \suml{j=1}{N_B}\beta^l_{ij}B_j.\nonumber 
\end{align}
\item[(iii)] Compute $\nabla_\vt L(u^*(\vt))$ at $\vt^l$ via \eqref{overallderivative}, 
\item[(iv)] Update parameters $\vt = \{\alpha, \beta\}$ by using a quasi-Newton's method, let $l = l+1$, and goto (ii).
\end{itemize}
\end{algorithm}
In \eqref{overallderivative}, we have collected all the necessary information to compute the 
gradients of the loss function with respect to the parameters $\vt$, 
so we can now employ gradient descent based algorithms, \eg, the steepest descent method, for optimization. 
Although this type of algorithm is very easy to implement, it is not efficient. 
In this paper, we turn to 
a more efficient non-linear optimization method - the Limited-memory Broyden-Fletcher-Goldfarb-Shanno (L-BFGS) 
quasi-Newton's method~\cite{BFGS}. 
We summarize our bi-level learning scheme in Algorithm~\ref{algo3.1}.

In our work, step (ii) in Algorithm~\ref{algo3.1} 
is completed using the L-BFGS algorithm, as this problem is smooth, to which L-BFGS is perfectly applicable. 
We solve this minimization problem to a very high accuracy with $\|\nabla_uE(u^*)\|_2 \leq 10^{-3}$ 
(gray-values in the range [0 255]), i.e., we use a more conservative convergence criterion in this inner loop 
than previous work \cite{SamuelFoE}. 
The training algorithm is terminated when the relative change of the loss 
is less than a tolerance, \eg, $tol = 10^{-5}$, a maximum number of iterations \eg, $maxiter = 500$ is reached 
or L-BFGS can not find a feasible step to decrease the loss. 

\section{Training experiments}\label{trainingexperiments}
\begin{figure}[t!]
\begin{center}
    {\includegraphics[width=0.48\textwidth]{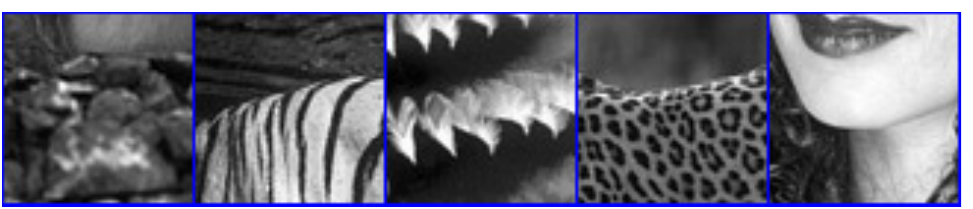}}\\\vspace{-0.2cm}
    {\includegraphics[width=0.48\textwidth]{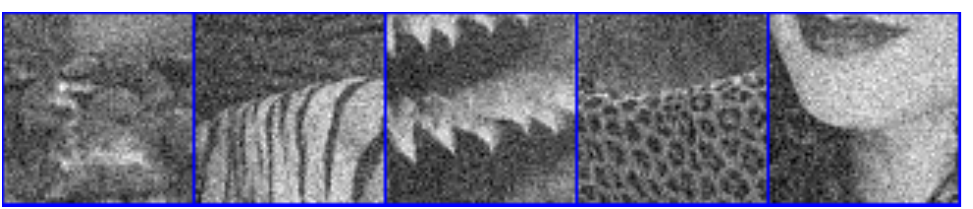}}
\end{center}
    \caption{Subset of the ground truth and the noisy data with noise level $\sigma=25$.}
\label{fig:patches}
\end{figure}

We conducted our training experiments using the training images from the BSDS300 image 
segmentation database~\cite{amfm_pami2011}. 
We used the whole 200 training images, and randomly sampled one $64 \times 64$ patch from each training image, 
resulting in a total of 200 training samples. 
We then generated the noisy versions by adding Gaussian noise with standard deviation 
${\sigma = 25}$. Figure~\ref{fig:patches} 
shows an exemplary subset of the training data together with the noisy version. 

In order to evaluate the performance of the learned analysis operators, we applied them to the image 
denoising experiment over a validation dataset consisting of 68 images from Berkeley database 
\cite{amfm_pami2011}. This is a common denoising test dataset for natural images, which was selected by 
Roth and Black \cite{RothFOE2009}. The performance of an image denoising algorithm varies greatly 
for different image contents. We  
therefore consider the average performance over the whole test dataset as performance measure. 

\subsection{Penalty functions}
So far we have not designated the penalty function $\phi$ used in our training model. 
In this paper, we consider three 
penalty functions with different properties: (1) the $\ell_1$ norm, which is a well-known 
convex sparsity promoting function and 
has been successfully applied to a number of problems in image restoration \cite{DonohoET06}, 
(2) the log-sum penalty 
suggested in \cite{Enhancing08}, $\text{log}(1+|z|)$, which is a non-convex function and can 
enhance sparsity and (3) the smooth 
non-convex function $\text{log}(1+z^2)$, which is derived from the student-t distribution and 
has been employed as 
the penalty function for sparse representation \cite{TehWOH03}. As our training model needs 
differentiable penalty functions, we 
have to use a small parameter $\eps$ to regularize the absolute function $|z|$. The penalty 
functions and their associated derivatives are given by
\begin{equation}\label{gx}
\begin{aligned}
\begin{cases}
\phi(z) = \sqrt{z^2 + \eps^2}\\
\phi'(z) = z/\sqrt{z^2 + \eps^2}\\
\phi''(z) = \eps^2/(z^2 + \eps^2)^{3/2} 
\end{cases}
&
\begin{cases}
\phi(z) = \text{log}(1 + z^2)\\
\phi'(z) = 2z/{(1 + z^2)}\\
\phi''(z) = 2(1 - z^2)/{(1+z^2)^2}, 
\end{cases}
\end{aligned}
\end{equation}
and
\begin{equation*}
\begin{cases}
\phi(z) = \text{log}(1 - \eps + \sqrt{z^2 + \eps^2})\\
\phi'(z) = \frac{z}{\sqrt{z^2 + \eps^2}(1 - \eps + \sqrt{z^2 + \eps^2})}\\
\phi''(z) = \frac{\eps^2(1-\eps) + (\eps^2-z^2)\sqrt{z^2 + \eps^2}}{(z^2 + \eps^2)^{3/2}
(1 - \eps + \sqrt{z^2 + \eps^2})^2}. 
\end{cases}
\end{equation*}
\subsection{Training experiments}
We focused training on filters of dimension $7 \times 7$, since 
our approach allowed us to train larger filters than those trained  
in~\cite{GaoCVPR2010,GaoDAGM2012}, and normally larger filters can involve more information of the 
neighborhood. First, we conducted a preliminary training experiment based on the 
penalty function $\text{log}(1+z^2)$. 
We intended to learn an analysis operator $A \in \R^{48 \times 49}$, i.e, 48 filters with 
dimension $7 \times 7$, and each filter 
is expressed as a linear combination of the DCT-7 basis. 
In principle, we can use any basis such as the identity, PCA or ICA basis; 
however as described below, we need zero-mean filters, which is guaranteed by the DCT filters after excluding 
the filter with constant entries. 

For the preliminary experiment, 
we initialized the analysis operator using 48 random filters having unified norms and weights, which are 
0.01 and 1, respectively. 
Finally training result shows that all the coefficients with respect to 
the first atom of DCT-7 (an atom with constant entries) are approximately equal to zero, 
implying that the first atom isn't necessary to construct the filters. 
Therefore the learned filters are undoubtedly zero-mean because 
all the remaining atoms are zero-mean; 
this makes the analysis prior based model~\eqref{analysismodel} illumination invariant. 
This result is coherent with the findings in 
the work \cite{Huang1999_Statistics} that meaningful filters should be zero-mean. 
Then we explicitly exclude the first atom in DCT-7 to speed up the training process 
for the following experiments.

\begin{figure}[t!]
  \begin{center}
{\includegraphics[width=0.45\textwidth]{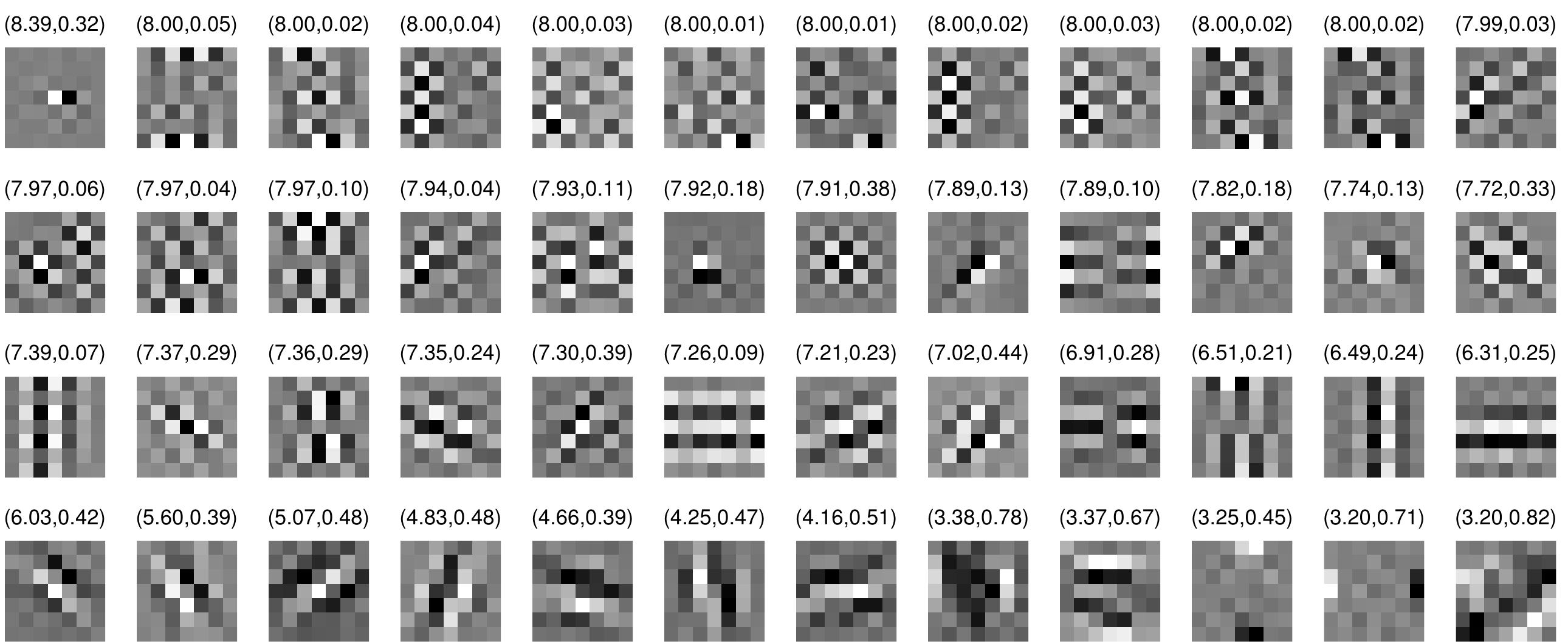}}    
    \caption{48 filters of size $7 \times 7$ learned by using the penalty function $\text{log}(1+z^2)$. 
Each filter is shown with the corresponding norm and weight. The 
first number in the bracket is the weight $\alpha_i$ and the second one is the norm of the filter.}
\label{ST48filters}
  \end{center}
\end{figure}

We then conducted training experiments based on three different penalty functions. In this paper, the regularization parameter 
$\eps$ in \eqref{gx} was set to $\eps = 10^{-2}$. 
{Smaller $\eps$ implies a better fitting to the absolute function, but it makes the lower-level problem harder to solve and the 
training algorithm fail.} 
Just like in the preliminary experiment, we also learned 48 filters. 
We initialized the filters using the modified DCT-7 basis with unified norms and weights. 
The optimal analysis operator learned by using the penalty function $\text{log}(1+z^2)$ is shown in Figure~\ref{ST48filters}. 
The final loss function values (normalized by the number of training images) of these three experiments 
are presented in Table~\ref{learning} (first three columns), together with the average denoising PSNR 
results based on 68 test images with $\sigma = 25$ Gaussian noise. 

\begin{table*}[t!]
\begin{center}
\begin{tabular}{|l|c |c |c ||c |c| c| c| c|}
\hline
\multicolumn{9}{c}{Final training results and the average denoising PSNR results on 68 test images}\\
\hline \hline
Penalty function  & $|z|$ & $\text{log}(1+|z|)$ & $\text{log}(1+z^2)$ & $\text{log}(1+z^2)$ 
& $\text{log}(1+z^2)$ & $\text{log}(1+z^2)$ & $\text{log}(1+z^2)$ & $\text{log}(1+z^2)$\\
\cline{1-9}
Filter size & $7 \times 7$ & $7 \times 7$ & $7 \times 7$ & $7 \times 7$ & $9 \times 9$
& direct DCT-7 & $5 \times 5$ & $3 \times 3$\\
\cline{1-9}
Number of filters & 48 & 48 & 48 & 98 & 80 & 48 & 24 & 8\\
\cline{1-9}
Final loss value &440,350 & 389,860& 388,053& 386,270 & 384,788 & 407,556 & 396,250 & 437,788\\
\cline{1-9}
average PSNR & 28.04 & 28.64 & 28.66 & 28.68 & 28.70 & 28.47 & 28.56 & 28.13\\
\cline{1-9}
\end{tabular}
\end{center}
\caption{Summary of the final training loss values for different model capacities and 
the corresponding average denoising PSNR results based on 
68 test images with $\sigma = 25$ Gaussian noise.}
\label{learning}
\end{table*}

As shown in Figure~\ref{ST48filters}, the learned filters present some special structures. 
We can find high-frequency filters as well as  derivative 
filters including the first derivatives along different directions, the second and 
the third derivatives. 
These filters make the analysis prior based model \eqref{analysismodel} a higher-order 
model which is able to capture the structures in natural images 
that cannot be captured by using only the first derivatives as in the total 
variation based methods. 

In our training model, the size and the number of filters are free parameters; thus 
we can train filters of various sizes and numbers. Our current implementation is 
unoptimized Matlab code. The training time for 48 filters of size $7 \times 7$ was approximately 24 hours on a server 
(Intel X5675, 3.07GHz), 98 filters of size $7 \times 7$ took about 80 hours. 
However, the training time for larger filter size $9 \times 9$ was much longer; it took about 20 days. 
Fortunately, the training procedure is off-line; thus the training time does not matter too much in practice. 
Demo training code can be downloaded from our homepage \footnote{\url{www.gpu4vision.org}}.
\subsection{The influence of the  penalty function}
\begin{figure}[t!]
  \begin{center}
    {\includegraphics[width=0.3\textwidth]{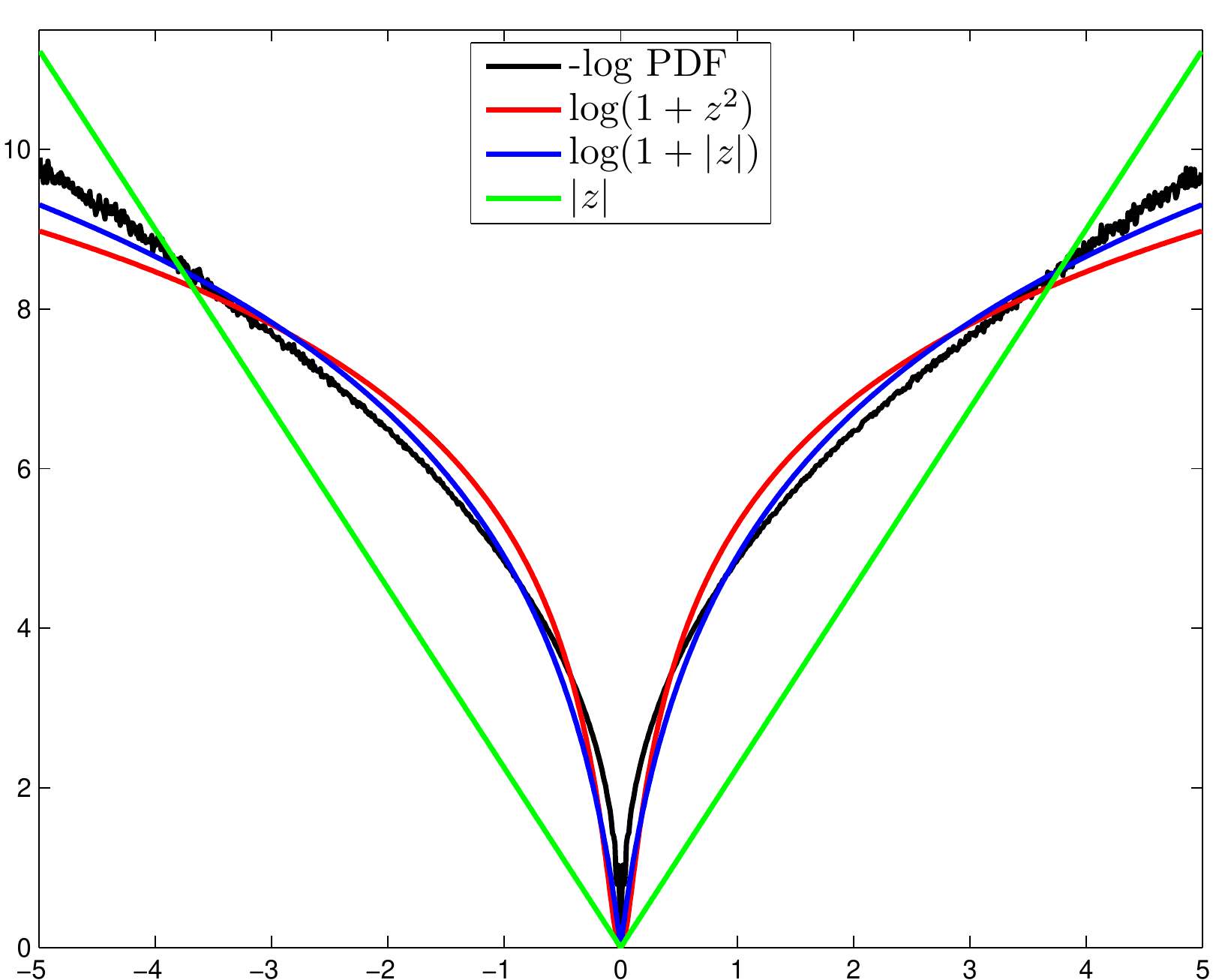}}
    \caption{Negative log probability density function (PDF) of the filter response of a learned $7 \times 7$ filter 
applied to natural images. Note that non-convex functions $\text{log}(1+|z|)$ and 
$\text{log}(1+z^2)$ provide much better fits to the heavy tailed shape of the true density function, 
compared to the convex function $|z|$.}
\label{pdf}
  \end{center}
\end{figure}
\begin{figure}[t!]
  \begin{center}
    {\includegraphics[width=0.3\textwidth]{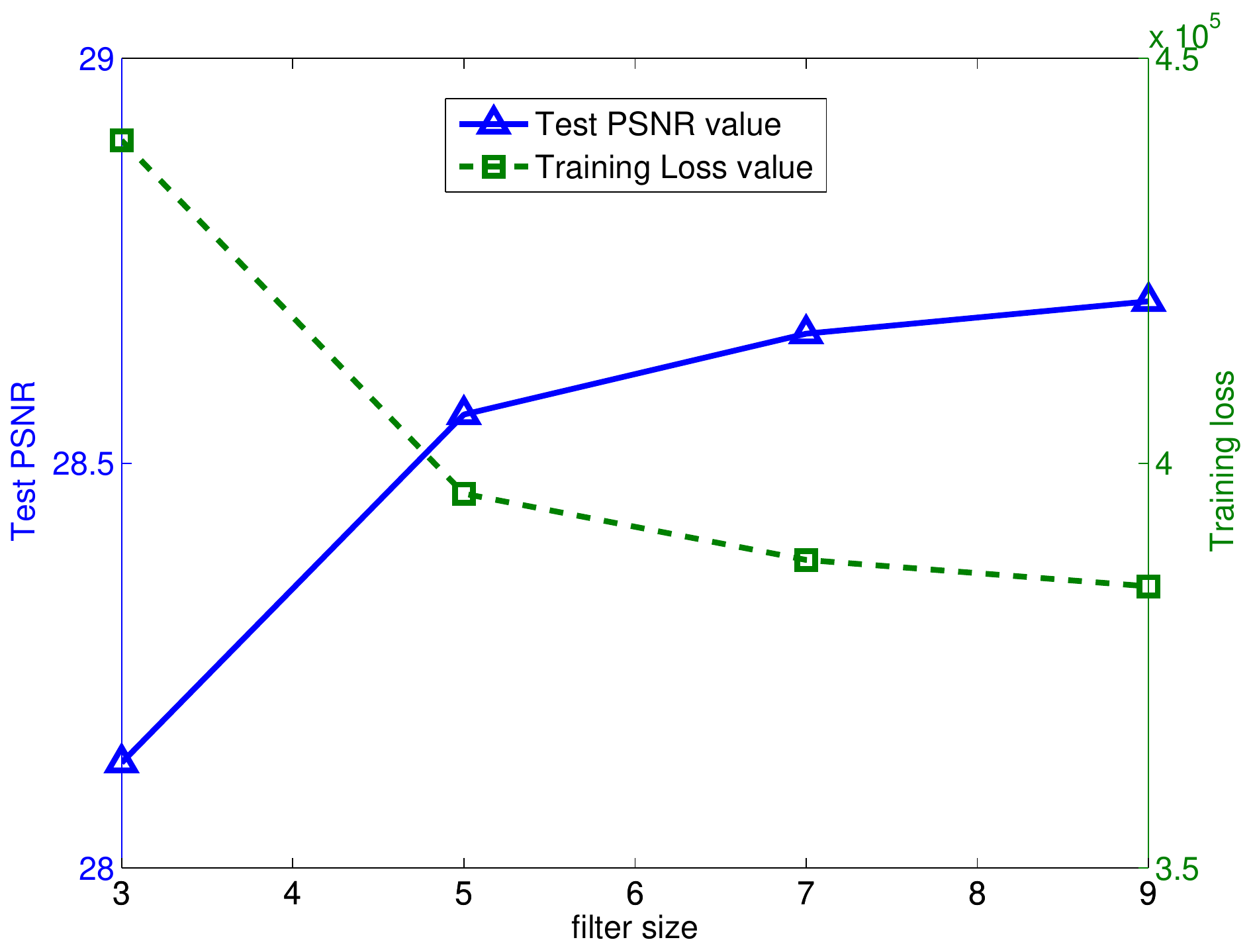}}
    \caption{Performance curves (test PSNR value and training loss value) \textit{vs.} the filter size. One can see that 
larger filter size can yield mprovements. }
\label{filtersize}
  \end{center}
\end{figure}
From Table~\ref{learning}, we can see that the results obtained by two non-convex penalty 
functions, $\text{log}(1+|z|)$ and 
$\text{log}(1+z^2)$ are very similar; however, there is a great improvement compared 
with the convex $\ell_1$ penalty function. The reason is as follows. 
It is well known that the probability density function (PDF) of the responses
of zero-mean linear filters (e.g., DCT filters) applied to natural images exhibit heavy tailed distributions~
\cite{Huang1999_Statistics}. 
Figure~\ref{pdf} shows the negative log PDF of the first filter in Figure~\ref{ST48filters} 
applied to natural images together with different model fits. 
We can clearly see that two non-convex functions, $\text{log}(1+|z|)$ and 
$\text{log}(1+z^2)$ both provide an almost perfect fit to the heavy tailed shape of the true density 
function. The convex function $|z|$ presents a much worse fitting, however.
Therefore, a suitable penalty function is crucial for the analysis-prior based model. In general, in order 
to model the heavy tailed shape of the true PDF, a non-convex function is required. 

In order to further investigate how important the non-convex penalty function is for the analysis prior based model, 
we considered an analysis model consisting of 48 fixed and predefined filters (DCT-7 filters excluding the filter with uniform entries) 
and making use of the $\text{log}(1+z^2)$ penalty function. We only optimized the norm and weight of each filter by using 
our bi-level training algorithm. The training loss value and the denoising test result of this model are shown in Table~\ref{learning} 
(the sixth column entitled ``direct DCT-7''). The image denoising test result is surprisingly good, even though this 
analysis model only utilizes a predefined analysis operator DCT-7. We will see in Table~\ref{comparison} of 
Section~\ref{applicationresults} that 
the performance of this model is already on par with the currently best analysis operator 
learning model - GOAL \cite{HaweAnalysisLearning}, which involves much more carefully trained filters. 
The success of this model lies in the non-convex penalty 
function $\text{log}(1+z^2)$. 

\subsection{The influence of the number of filters}
In previous work of analysis operator learning {\cite{RubinsteinKSVD,
YahoobiAnalysisLearning,HaweAnalysisLearning}}, the  authors were interested in the over-complete case, where the 
number of filters is larger than the dimension of filters. Clearly our learned analysis operator in Figure~\ref{ST48filters} is 
under-complete. In order to investigate the influence of the over-complete property, we also conducted a training experiment 
for the over-complete case ($A \in \R^{98 \times 49}$) based on the penalty function $\text{log}(1+z^2)$. 
We initialized the analysis operator $A$ using 98 random zero-mean filters. 
The performance of this over-complete case is presented in Table~\ref{learning}.

From Table~\ref{learning}, one can see that the improvement achieved by over-complete analysis operator 
is marginal. Therefore for the analysis model, under-complete operators already work sufficiently well. 
An increase of the number of filters can not bring large improvements. 
\subsection{The influence of filter size}
Intuitively the size of filters should be an important factor for the analysis model. 
In order to investigate the influence of filter size, we conducted training experiments for several different 
analysis models, where the filter size varies from $3 \times 3$ to $9 \times 9$. The training and evaluation results of 
these models are presented in Table~\ref{learning} and Figure~\ref{filtersize}. 

One can see that increasing the filter size yields some improvements. However, the performance 
is close to saturation when the filter size is increasing to $7 \times 7$. The improvement brought by 
increasing the filter size to $9 \times 9$ is negligible. This implies that we can not expect large 
improvements by increasing the filter size to $11 \times 11$ or larger. 
\subsection{The robustness of our training scheme}
As our training model \eqref{learningmodel} is a non-convex optimization problem, we can only find local minima. 
Thus a natural question about the initialization arises. We did have experiments for different initializations, such as 
random initialization. The final learned analysis operators are surely different, but all of them have almost the same training 
loss, which is the goal of our optimization problem. In addition, these operators perform similarly in evaluation experiments. 

Another issue about the robustness of our training scheme is the influence of the training dataset. 
Since the training patches were randomly selected, we could run the training experiment
multiple times by using a different training dataset. Finally, we found that the deviation of test PSNR values based on 
68 test images is within 0.02dB, which is negligible. 

\section{Application results using learned operators}\label{applicationresults}
An important question for a learned prior model is how well it generalizes. To
evaluate this, we directly applied our learned analysis operators, which where trained for
the image denoising task, to various image restoration problems such as image
deconvolution, inpainting and super-resolution, as well as denoising. 
To start with, we first express the image restoration model by using our learned analysis operator, 
which is formulated as:
\begin{align}\label{generalmodel}
\arg\min\limits_{\imgX}
E(\imgX) =  \suml{i=1}{n} \alpha_i \phi(\cA_i \imgX) + 
\frac{\lambda}{2}\|K\imgX-\imgY\|_2^2,
\end{align}
where $K$ is a linear operator which depends on the respective application 
to be handled. 

To solve the minimization problem \eqref{generalmodel}, for convex $\ell_1$ based model, 
we used the first-order primal-dual algorithm proposed in \cite{pdpock} 
with the preconditioning technique described in \cite{pockprepd}; 
for non-convex penalty function based model, we employed L-BFGS for optimization. 
For L-BFGS, we need to calculate the gradient $\nabla_uE$, which includes constructing the highly sparse 
large matrix $\cA_i$ and its transpose $\cA_i^T$. This is quite time consuming in practice. In order to speed up 
the inference algorithm, we turn to a filtering technique as we know that $\cA_iu$ is equivalent to the filtering operation ($A_i * u$). 
Therefore, the gradient $\nabla_uE$ is given by
\begin{equation}\label{gradient}
\nabla_uE = 
\suml{i=1}{n} \alpha_i A_i^- * \phi'(A_i * \imgX) + 
\lambda K^T(K\imgX-\imgY), 
\end{equation}
where $A_i * \imgX$ denotes the convolution of image $\imgX$ with filter $A_i$, and $A_i^-$ denotes the filter obtained 
by mirroring $A_i$ around its center pixel 
(in practice, we need to carefully handle the boundaries as we need to ensure that the filters $A_i^-$ correspond
to $\cA_i^T$). 

We provide Matlab demo code for training and denoising with penalty function $\text{log}(1+z^2)$ 
on our homepage \textit{\url{www.gpu4vision.org}}.
\subsection{Image denoising results}\label{denoising}
We first apply the analysis model based on our learned operators to the image denoising problem. 
In the case of image denoising, $K$ is simply the identity matrix, i.e., $K = \cI$. 
Since the image denoising performance of one method varies greatly for 
different image contents, in order to make a fair 
comparison, we conducted denoising experiments over a standard test dataset - 
68 Berkeley test images identified by Roth and Black 
\cite{RothFOE2009}. We used exactly the same noisy version of each test 
image for different methods and different test images were added with distinct noise realizations. 
All results were computed per image and then averaged over the test dataset. 

We considered image denoising for 
various noise levels $\sigma = \{15,25,50\}$. 
For noise levels other than $\sigma = 25$, we need to tune the parameter $\lambda$ in \eqref{generalmodel}. 
An empirical choice of $\lambda$ is: $\sigma = 15$, 
$\lambda = 25/\sigma\times1.15$; $\sigma = 50$, $\lambda = 25/\sigma\times0.8$. 
\begin{figure}[t!]
  \begin{center}
    \subfigure[noisy (20.17)]{\includegraphics[width=0.242\textwidth]{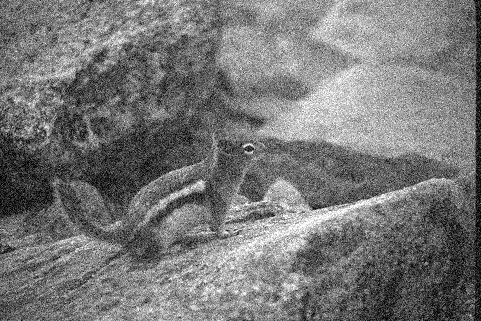}}\hfill
    \subfigure[noisy (20.17)]{\includegraphics[width=0.242\textwidth]{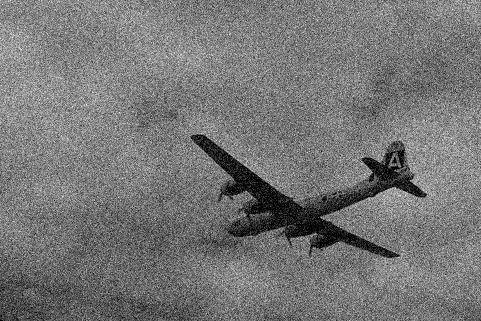}}\\
    \subfigure[$\text{log}(1+z^2)$ (29.31)]{\includegraphics[width=0.242\textwidth]{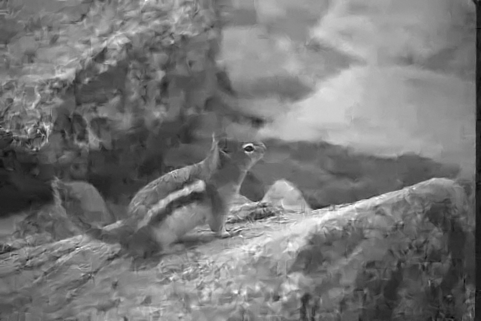}}\hfill
    \subfigure[$\text{log}(1+z^2)$ (36.84)]{\includegraphics[width=0.242\textwidth]{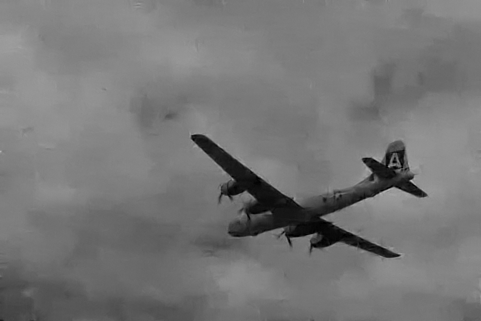}}\\
    \subfigure[$\text{log}(1+|z|)$ (29.34)]{\includegraphics[width=0.242\textwidth]{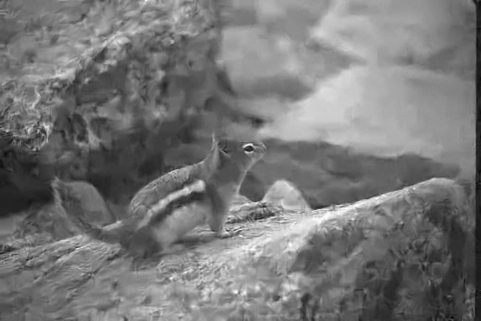}}\hfill
    \subfigure[$\text{log}(1+|z|)$ (36.40)]{\includegraphics[width=0.242\textwidth]{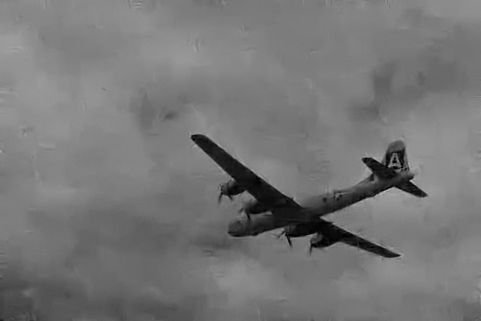}}\\
    \subfigure[$|z|$ (29.03)]{\includegraphics[width=0.242\textwidth]{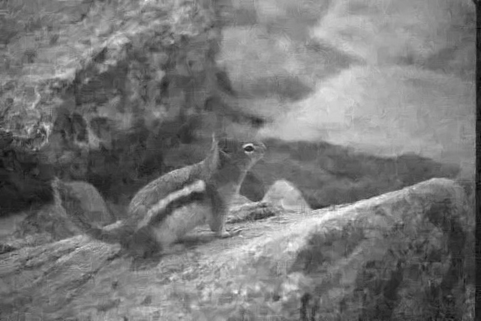}}\hfill
    \subfigure[$|z|$ (34.24)]{\includegraphics[width=0.242\textwidth]{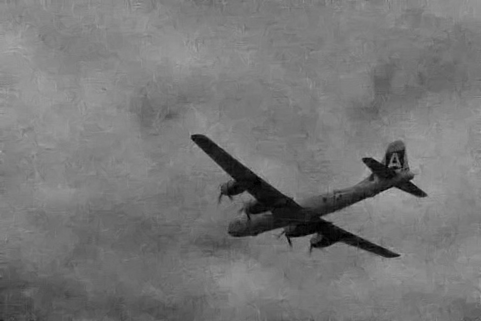}}
    \caption{Comparison of denoising results obtained by three different penalty functions for noise level $\sigma=25$. 
The numbers shown in the brackets refer to PSNR values with respect to the clean images. 
}\label{denoisingexamples}
  \end{center}
\end{figure}
\begin{table*}[t!]
\begin{center}
\begin{tabular}{|l|p{0.65cm}|p{0.5cm}|p{0.65cm}|p{0.65cm}|p{0.65cm}|p{0.65cm}||c|c|c|c|c|c|}
\hline
$\sigma$ & KSVD & FoE & GOAL & BM3D & LSSC & EPLL & \tabincell{c}{log$(1+z^2)$\\$7 \times 7$, 48} 
& \tabincell{c}{log$(1+|z|)$\\$7 \times 7$, 48} & \tabincell{c}{$|z|$ \\$7 \times 7$, 48} 
& \tabincell{c}{log$(1+z^2)$\\$7 \times 7$, 98} & \tabincell{c}{log$(1+z^2)$\\$9 \times 9$, 80}
& \tabincell{c}{log$(1+z^2)$\\ direct DCT-7}\\
\hline\hline
15 & 30.87 & 30.99 & 31.03 & 31.08 & \textbf{31.27} & 31.19 & 31.18 & 31.18 & 30.45 & \textbf{31.22} & \textbf{31.22} & 30.92\\
\cline{1-13}
25 & 28.28 & 28.40 & 28.45 & 28.56 & \textbf{28.70} & \textbf{28.68} & \textbf{28.66}  &{28.64} & 28.04 & \textbf{28.68} & \textbf{28.70} &{28.47} \\
\cline{1-13}
50 & 25.17 & 25.35 & 25.44 & 25.62 & \textbf{25.72} & \textbf{25.67} & \textbf{25.70} & 25.58 & 25.12 & \textbf{25.71} & \textbf{25.76} & 25.58 \\
\cline{1-13}
\end{tabular}
\end{center}
\caption{Summary of denoising results (average PSNR values). 
We highlighted the state-of-the-art results.}
\label{comparison}
\end{table*}
\subsubsection{Comparison of three different penalty functions}
Table~\ref{comparison} shows the summary of denoising results achieved by different penalty functions. 
One can clearly see that 
two non-convex penalty functions lead to similarly good results and they significantly outperform the results 
of the convex function $|z|$. 
In addition, we can also see that the over-complete operator can not improve the performance too much 
and larger filters ($9 \times 9$) can only achieve slightly better performance. Both of these two models are more time 
consuming than the model with 48 filters for inference; therefore, 
the analysis model based on 48 filters of size $7 \times 7$ offers the best trade-off between computational cost and performance. 
In the following experiments, 
we only consider the model of 48 filters and the penalty function $\text{log}(1+z^2)$. We prefer the penalty function $\text{log}(1+z^2)$, 
since it is completely smooth, making the corresponding minimization problem easier to solve. 
We present two denoising examples 
obtained by three different penalty functions in Figure~\ref{denoisingexamples}. 

\subsubsection{Comparison to other analysis models}
Our learned model is an analysis prior-based model, but can also be viewed as a MRF-based system. In order to rank our model among 
other analysis models, we compared its performance with existing analysis models, including typical FoE models 
\cite{RothFOE2009,GaoCVPR2010,SamuelFoE,Barbu2009,DomkeAISTATS2012,GaoDAGM2012}
and the currently published best analysis operator model - GOAL \cite{HaweAnalysisLearning}. For the GOAL method, 
we made use of the $l_{0.4}$-norm penalty function $|z|^{0.4}$, together with 
the learned analysis operator $\Omega \in \R^{98 \times 49}$ provided by the authors. 
We also utilized the L-BFGS algorithm to solve the 
corresponding minimization problem. 
We present image denoising results of all approaches 
over 68 test images with noise level $\sigma = 25$ in Table~\ref{mrfs}. 
One can see that our model based on the penalty function $\text{log}(1+z^2)$ (48 learned filters, $7 \times 7$) has 
achieved the best performance among all the related approaches. The comparison with the best FoE 
\cite{GaoDAGM2012} model and 
the latest analysis model GOAL for other noise levels is shown in Table~\ref{comparison}. For all the noise levels, our 
trained analysis model outperforms both of them significantly. 

An interesting result in Table~\ref{comparison} is that the performance of the direct DCT-7 model, which 
only utilizes a predefined analysis operator DCT-7 (48 filters of size $7 \times 7$), 
is already on par with the GOAL model, 
which involves much more carefully trained filters (98 filters of size $7 \times 7$). 
{For this direct DCT-7 model, we used the $\text{log}(1+z^2)$ penalty function, and only 
optimized the norms and weights of the filters using our bi-level training algorithm.} 
This result demonstrates the 
importance of non-convex penalty functions and the effectiveness of our bi-level training scheme. 

As the analysis operator of GOAL model is trained using a patch-based model, 
we can also use it in the manner of patch-averaging to conduct image denoising like K-SVD \cite{KSVDdenoising2006}. 
We embedded the learned analysis operator $\Omega$ into the patch-based analysis model \eqref{analysispatch}, and 
used it to denoise each patch extracted from an image. We also considered overlapped windows and averaged the results 
in the overlapping regions to form the final denoised image. As expected, we got inferior results (average PSNR 28.25 
over 68 test images) to the model formulated under the FoE framework (average PSNR 28.45). 
\begin{table}[t!]
\begin{center}
\begin{tabular}{|L{1.2cm}|l|l|l|}
\hline
model & potential & training & PSNR\\
\hline\hline
$5 \times 5$ FoE& ST\&GLP. & contrastive divergence & 27.77\cite{RothFOE2009} \\
\cline{1-4}
$3 \times 3$ FoE& GSMs & contrastive divergence & 27.95\cite{GaoCVPR2010} \\
\cline{1-4}
$5 \times 5$ FoE&  GSMs & persistent contrastive divergence& 28.40\cite{GaoDAGM2012}\\
\cline{1-4}
$5 \times 5$ FoE&  ST & bi-level (truncated optimization) & 28.24\cite{Barbu2009}\\
\cline{1-4}
$5 \times 5$ FoE&  ST & bi-level (truncated optimization) & 28.39\cite
{DomkeAISTATS2012}\\
\cline{1-4}
$5 \times 5$ FoE&  ST & bi-level (implicit differentiation) & 27.86\cite{SamuelFoE}\\
\cline{1-4}
$7 \times 7$ \newline GOAL&  GLP & geometric conjugate gradient & 28.45\cite{HaweAnalysisLearning}\\
\cline{1-4}
$7 \times 7$ FoE&  ST & bi-level (implicit differentiation) & \textbf{28.66}\\
\cline{1-4}
\hline
\end{tabular}
\end{center}
\caption{Summary of various analysis models and the average denoising results on 68 test images 
with $\sigma = 25$. We highlighted our result, also the best one.}
\label{mrfs}
\end{table}

\begin{figure}[t!]
\vspace*{-0.75cm}
    \includegraphics[width=0.4\textwidth]{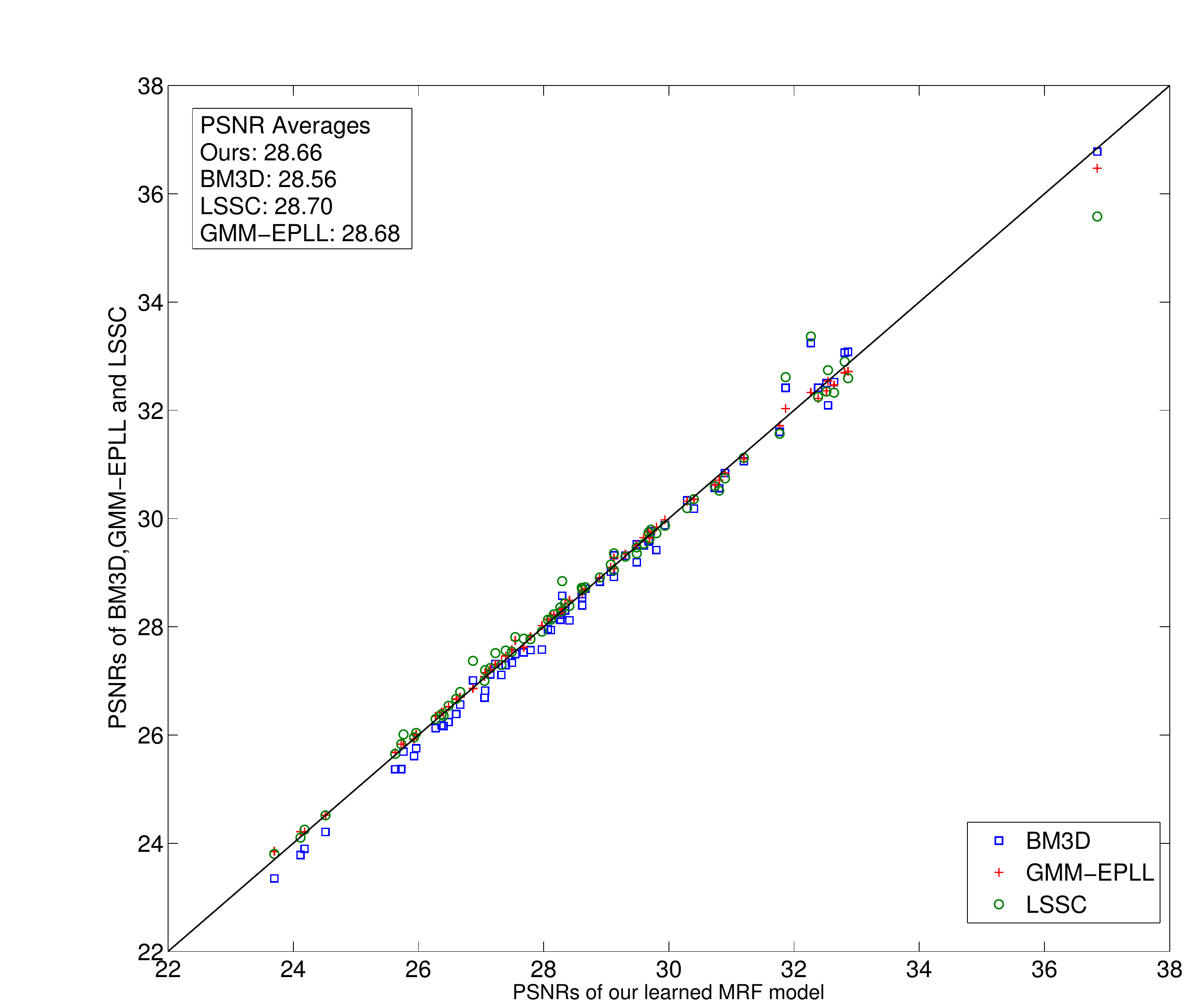}
    \caption{Scatter plot of the PSNRs over 68 Berkeley images produced by 
our learned log($1+z^2$)-based analysis model, 
BM3D, GMM-EPLL and LSSC. A point above the line $y = x$ means a better performance than our model.}\label{fig:comparison}
\end{figure}

\subsubsection{Comparison to state-of-the-art methods}
In order to evaluate how well our analysis models work for the denoising task, we compared their performance with leading image 
denoising methods, including three state-of-the-art methods: (1) BM3D~\cite{BM3D}; (2) LSSC~\cite{LSSC}; 
(3) GMM-EPLL~\cite{EPLL} along with three leading generic methods: (4) a MRF-based approach, FoE~\cite{GaoDAGM2012}; 
(5) a synthesis sparse representation based method, KSVD~\cite{KSVDdenoising2006} trained on natural image patches; and (6) 
the currently published best analysis operator learning method, GOAL \cite{HaweAnalysisLearning}. 
All implementations were downloaded from the corresponding authors' homepages. 
We conducted denoising experiments over 68 Berkeley test images with various noise levels 
$\sigma = \{15, 25, 50\}$. 
All results were computed per image and then averaged over the number of images. 

\begin{figure}[t!]
\vspace*{-0.3cm}
  \begin{center}
    \subfigure[clean image]{\includegraphics[width=0.161\textwidth]{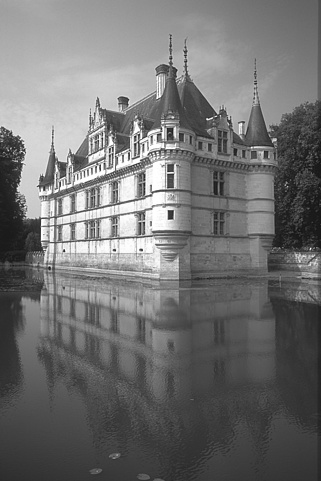}}\hfill
    \subfigure[$\sigma = 25$ (20.17)]{\includegraphics[width=0.161\textwidth]{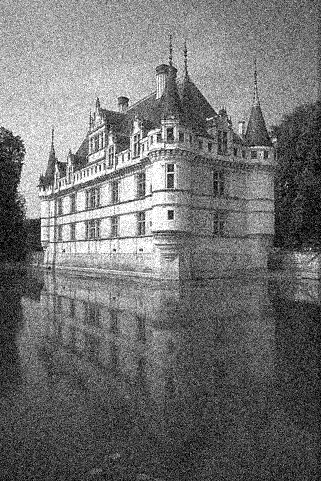}}\hfill
    \subfigure[KSVD (29.13)]{\includegraphics[width=0.161\textwidth]{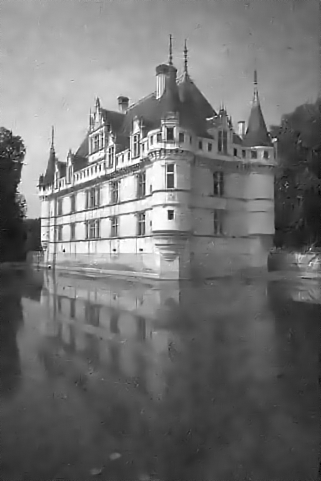}}\\
    \subfigure[FoE (29.15)]{\includegraphics[width=0.161\textwidth]{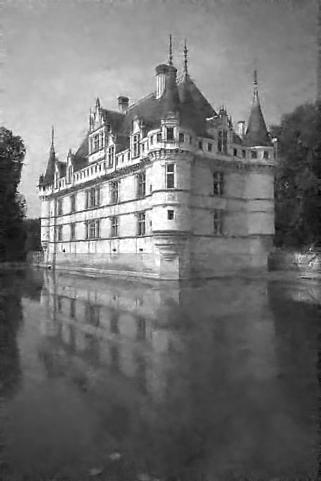}}\hfill
    \subfigure[BM3D (29.52)]{\includegraphics[width=0.161\textwidth]{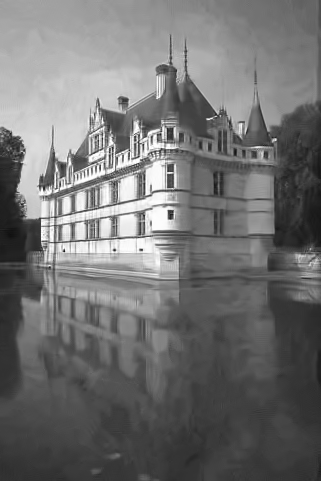}}\hfill
    \subfigure[LSSC (29.47)]{\includegraphics[width=0.161\textwidth]{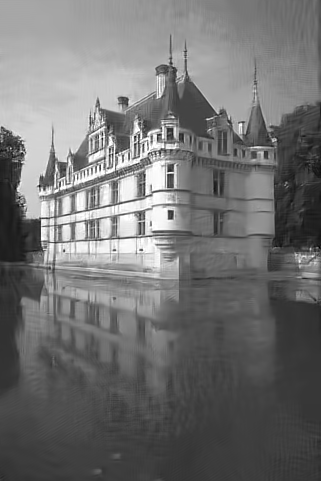}}\\
    \subfigure[GMM-EPLL (29.50)]{\includegraphics[width=0.161\textwidth]{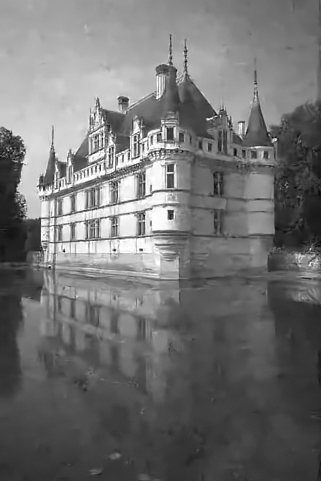}}\hfill    
    \subfigure[GOAL (29.30)]{\includegraphics[width=0.161\textwidth]{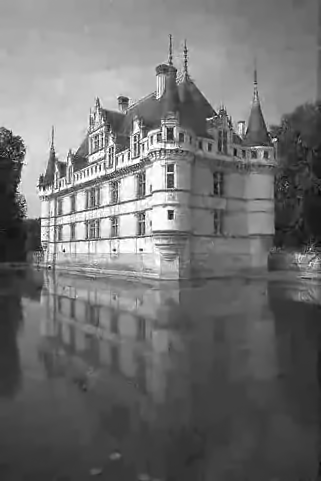}}\hfill
    \subfigure[$\text{log}(1+z^2)$ (29.48)]{\includegraphics[width=0.161\textwidth]{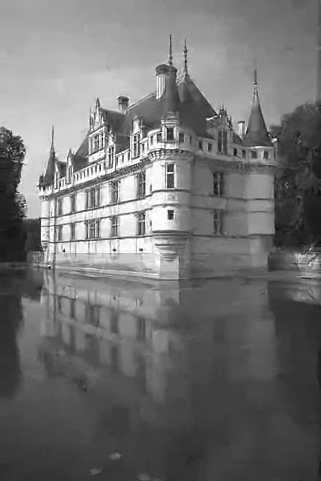}}\\
    \caption{The images $(c)\sim(i)$ present results achieved by 
the compared methods for denoising ``water-castle'' image corrupted with $\sigma = 25$.}
\label{comparisonfigure}
  \end{center}
\end{figure}

Table~\ref{comparison} shows the summary of results. 
One can see that our trained model based on the penalty function $\text{log}(1+z^2)$ (48 learned filters, $7 \times 7$) 
outperforms three leading generic methods and is on par with three state-of-the-art methods for any noise level. 
To the best of our knowledge, this is the first time that a MRF model based on generic priors of natural images has 
achieved such clear state-of-the-art performance. 
Figure~\ref{fig:comparison} gives a detailed comparison between our learned analysis model 
and three state-of-the-art methods over 68 
test images for $\sigma = 25$. We can see that all the points surround the diagonal line ``$y = x$'' 
closely, i.e., all considered methods achieve very 
similar results. Therefore, it is clear that our learned analysis models based on non-convex penalty functions 
are state-of-the-art. We present an image denoising example of the considered methods in Figure~\ref{comparisonfigure}. 

Our model is well-suited to GPU parallel computation since it solely contains 
convolution of filters with an image. Our GPU implementation based on a
NVIDIA Geforce GTX 580 accelerates the inference procedure significantly; for a denoising task with $\sigma = 25$, 
typically it takes 0.87s for image size $512 \times 512$, 0.60s for $481 \times 321$ and 0.29s for $256 \times 256$, 
i.e., using our GPU based implementation, image denoising can be conducted in near real-time at 3.4fps 
for an $256 \times 256$ image sequence, with  
state-of-the-art performance. In Table~\ref{runningtime}, we show the 
average running time of the considered denoising methods on $481 \times 321$ images. 

Considering the speed and quality of our model, it is a perfect choice as a base method in the image restoration framework 
proposed in~\cite{ECCV2012RTF}, which leverages advantages of existing methods. 

\subsection{Single image super-resolution}
\begin{figure}[t!]
  \begin{center}
    \subfigure[Original image]{\includegraphics[width=0.22\textwidth]{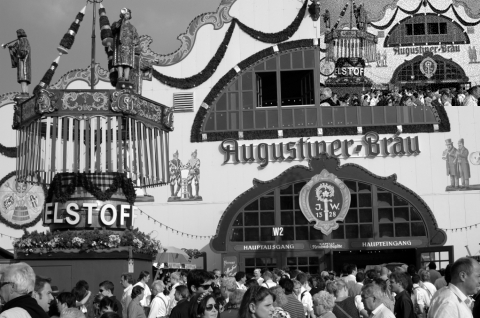}}
    \subfigure[Bicubic interpolation (21.63)]{\includegraphics[width=0.22\textwidth]{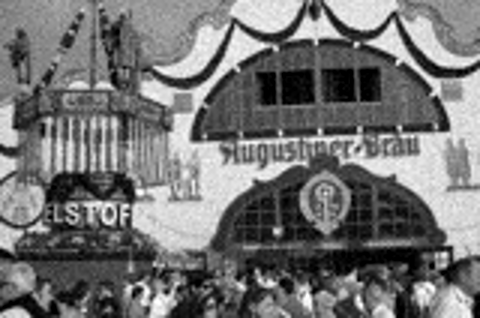}}\\
    \subfigure[GOAL \cite{HaweAnalysisLearning} (23.45)]{\includegraphics[width=0.22\textwidth]{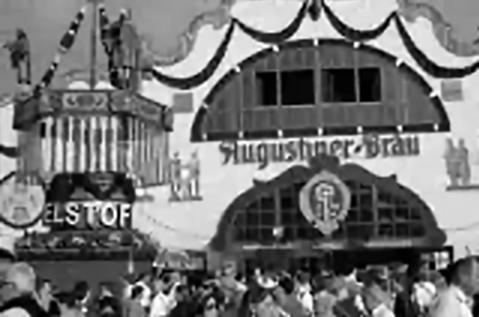}}
    \subfigure[Ours: log$(1+z^2)$ (23.50)]{\includegraphics[width=0.22\textwidth]{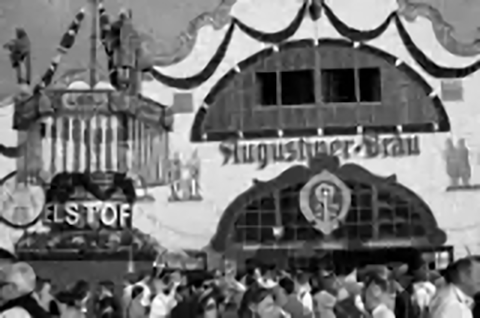}}
    \caption{Single image super-resolution results for magnifying a noisy low resolution image by a factor of 3. 
The low resolution image was degraded by Gaussian noise with $\sigma = 8$. 
The numbers shown in the brackets refer to 
PSNR values with respect to the clean image. }\label{eye}
  \end{center}
\end{figure}

For single image super-resolution, the linear operator $K$ is constructed by a decimation operator $\Phi$ and a 
blurring operator $B$, i.e., $K = \Phi B$. In order to perform a better comparison with the latest analysis 
model GOAL \cite{HaweAnalysisLearning}, we conducted the same single image super-resolution experiment. 
We artificially created a low resolution image by downsampling a ground-truth image by a 
factor of 3 using bicubic interpolation. Then the low resolution image was corrupted by Gaussian noise with $\sigma = 8$. 
We magnified the noisy low resolution image by the same factor using (a) bicubic interpolation, (b) GOAL method 
\cite{HaweAnalysisLearning}, 
(c) our learned analysis model based on penalty function log$(1+z^2)$ respectively. 
Figure~\ref{eye} shows the results for different methods. One can see that 
two analysis models present similar results, which are visually and quantitatively better 
than the bicubic method. 

\begin{table}[t!]
\begin{center}
\begin{tabular}{l|p{0.6cm}|p{0.5cm}|p{0.6cm}|c|c|c|c}
\cline{1-8}
 & KSVD & FoE & BM3D & LSSC & EPLL & GOAL & ours\\
\hline\hline
T(s) & 30 & 1600 & 4.3 & 700 & 99 & 112 & 12 (\textbf{0.6})\\
psnr & 28.28 & 28.40 & 28.56 & {28.70} & {28.68} & 28.45 & {28.66}\\
\cline{1-8}
\end{tabular}
\end{center}
\caption{Typical run time of the denoising methods for a $481 \times 321$ image ($\sigma = 25$) on a server 
(Intel X5675, 3.07GHz). The highlighted number is the run time of the GPU implementation.}
\label{runningtime}
\end{table}
\subsection{Non-blind image deconvolution}
\begin{figure*}[t!]
  \begin{center}
   \subfigure[Original image]{\includegraphics[width=0.19\textwidth]{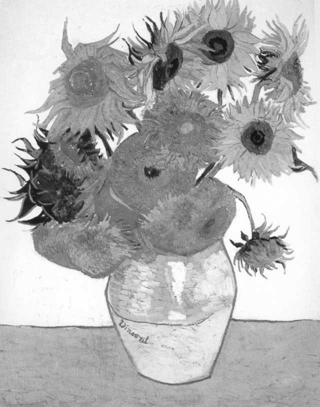}}\hfill
    \subfigure[Degraded image]{\includegraphics[width=0.19\textwidth]{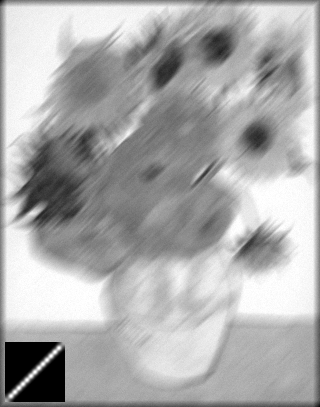}}\hfill
    \subfigure[GMM-EPLL (27.46)]{\includegraphics[width=0.19\textwidth]{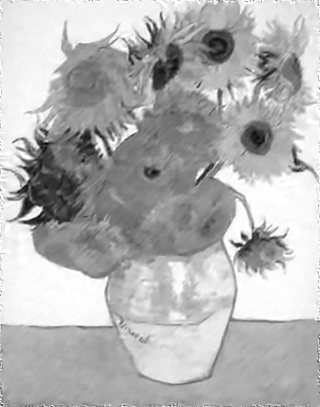}}\hfill
    \subfigure[GOAL (27.97)]{\includegraphics[width=0.19\textwidth]{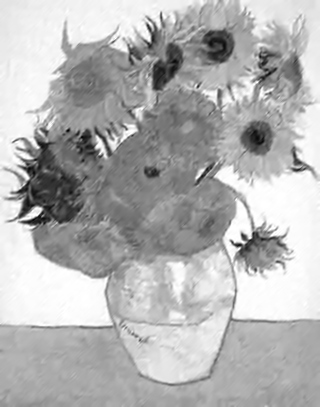}}\hfill
    \subfigure[log$(1+z^2)$ (28.26)]{\includegraphics[width=0.19\textwidth]{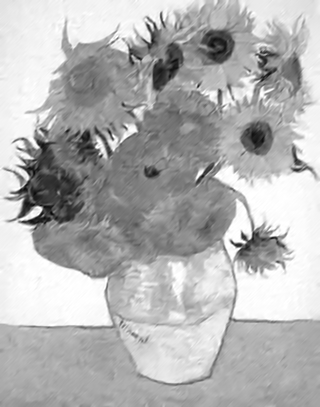}}    
    \caption{Motion deblurring results of our learned analysis model, the GMM-EPLL model~\cite{EPLL} and the GOAL model 
\cite{HaweAnalysisLearning}. The degraded image is generated using a motion blur of approximately 20 pixels and a Gaussian 
noise of $\sigma = 2.5$.}\label{sunflower}
  \end{center}
\end{figure*}

In the case of image deconvolution, $Ku$ can be written as a convolution, i.e., $Ku = k_K*u$, where $k_K$ is the convolution kernel. 
We considered a non-blind image deconvolution task, where a clean image was degraded by a motion blur of approximately 20 pixels 
and slight Gaussian noise with noise level $\sigma=2.5$. Again, we used our learned analysis models based on the log$(1+z^2)$ penalty function for image deblurring. 
In order to present a comparison, we also conducted deblurring by using the GMM-EPLL model~\cite{EPLL} and the latest 
analysis operator learning algorithm GOAL \cite{HaweAnalysisLearning}. For the GOAL method, we utilized the learned analysis operator 
$\Omega \in \R^{98 \times 49}$ provided by the authors. 
Figure~\ref{sunflower} shows the motion deblurring results achieved by our model, GMM-EPLL and GOAL, respectively. 
We can see that our learned analysis model yields better results in terms of PSNR. 

\subsection{Image inpainting}
Image inpainting is the process of filling in lost image data such that the resulting image is visually appealing. 
Typically, the positions of the pixels to be 
filled up are given. In our formulation, the linear operator $K$ is simply a sampling matrix, where each row 
contains exactly one entry equal to one. 
Its position indicates a pixel with given value. The parameter $\lambda$ corresponds to joint inpainting 
and denoising, and the choice 
$\lambda \rightarrow +\infty$ means pure inpainting. In our experiment since we assumed the test images 
are noise free, we empirically selected $\lambda = 10^3$. Due to space limitation, we 
only considered a classical image inpainting task here. 

\begin{figure*}[t!]
  \begin{center}
   \subfigure[90 \% missing pixels]{\includegraphics[width=0.242\textwidth]{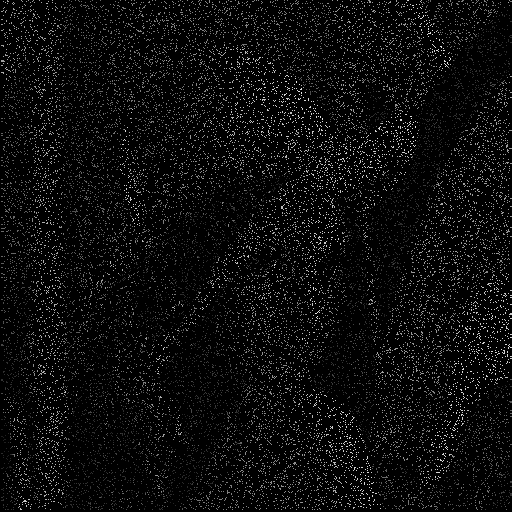}}\hfill    
    \subfigure[FoE (28.06)]{\includegraphics[width=0.242\textwidth]{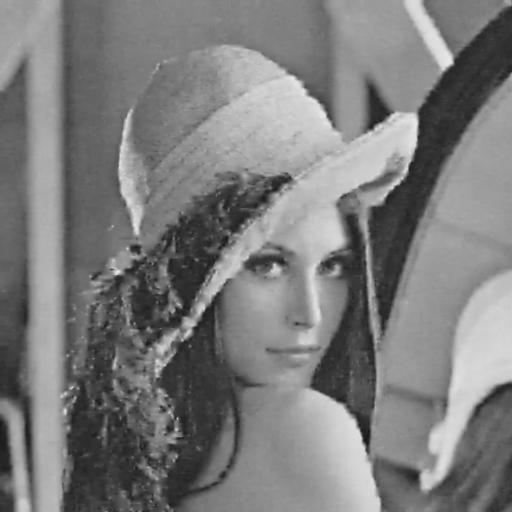}}\hfill
    \subfigure[GOAL (28.57)]{\includegraphics[width=0.242\textwidth]{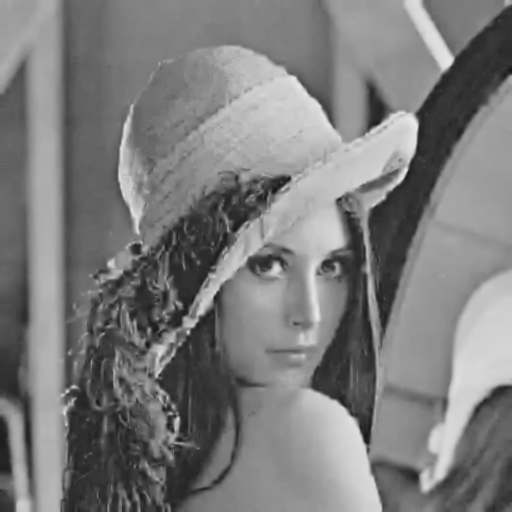}}\hfill
    \subfigure[log$(1+z^2)$ (28.62)]{\includegraphics[width=0.242\textwidth]{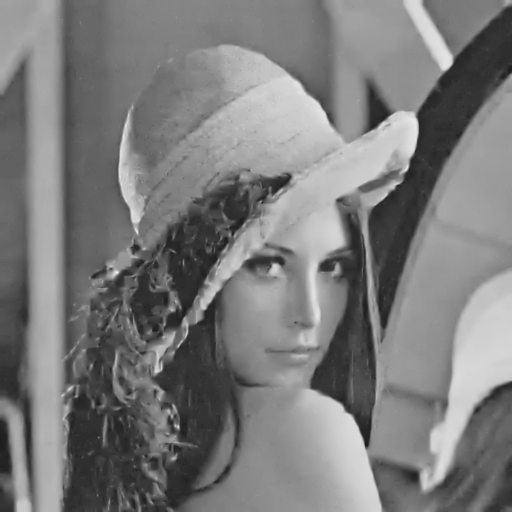}}\\    
    \caption{Recovery results of the destroyed \textquotedblleft Lena\textquotedblright~ image with 90\% loss pixels 
using our learned analysis model, a recent analysis model - GOAL~\cite{HaweAnalysisLearning} and the FoE model 
\cite{RothFOE2009}.}
\label{lena}
  \end{center}
\end{figure*}

We destroyed the ground-truth `` Lena'' image 
($512 \times 512$) 
artificially by masking 90\% of the entire pixels randomly as shown in Figure~\ref{lena}(a). 
Then we reconstructed the incomplete image using our learned analysis model - 
log($1+z^2$)-based model. In order to present a comparison, we also give the inpainting result 
of the GOAL model~\cite{HaweAnalysisLearning} and FoE model \cite{RothFOE2009}. 
From Figure~\ref{lena}, one can see that the result of our learned analysis model based on log($1+z^2$) penalty 
achieves equivalent results with respect to the GOAL model. 

\section{Conclusion and outlook}\label{conclusion}
In this paper, we have expressed our insights into the co-sparse analysis model. We 
propose to go beyond existing patch-based models, and to exploit the framework of FoE model to define 
a image prior over the entire image, rather than image patches. 
We have pointed out that the image based analysis model is equivalent to the FoE model. Starting from this conclusion, 
we have introduced a bi-level training approach for analysis operator learning, 
which is solved effectively with L-BFGS algorithm. 
By using our training framework, we have carefully investigated the effect of different aspects of the 
analysis prior model including the filter size, the number of filters and the penalty function. 

Since our training scheme directly optimizes the MAP-inference based analysis model, the learned model is an optimal 
MAP inference for image restoration problems. Numerical results have confirmed the good performance of 
our learned analysis model. For classic tasks such as 
image denoising, image deconvolution, image super-resolution and image inpainting,  
our learned analysis model has achieved strongly competitive performance with 
current state-of-the-art methods, and clearly outperforms existing analysis learning approaches. 

For future work, focusing on generic priors of natural images, we expect that our learned analysis model could be improved potentially 
in two aspects: (1) consider more flexible penalty function. 
In our current model, the penalty function is fixed to the same form for every filter. If we free the shape of the penalty function, 
our model will 
possess more freedom, which might increase the performance. A feasible way to consider alterable penalty function 
is to make use of the GSMs prior~\cite{GaoCVPR2010}. (2) make use of larger training dataset. Our training is conducted based 
on 200 training samples, which is only a very small part of the natural images. Consequently, the learned filters may 
over-fit on the training dataset. However, our current training scheme is not available for 
large training dataset, e.g., $\sim 10^6$, because it needs to solve the lower-level problem for each training sample. 
Feasible methods may include making use of stochastic optimization. 

\section{Acknowledgments}
The authors wish to thank Qi Gao (TU Darmstadt) for sharing details of the 
FoE model; Daniel Zoran (Hebrew University of Jerusalem) for sharing testing details for the GMM-EPLL algorithm; 
Simon Hawe (TU Munich) for useful discussion about analysis operator learning. 

\bibliographystyle{ieee}
\bibliography{bilevel_learning}
\vspace*{-1cm}
\begin{IEEEbiography}[{\includegraphics[width=1in,height=1.25in,clip,keepaspectratio]
{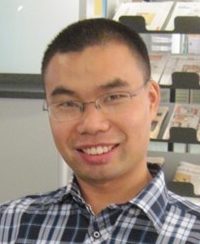}}]{Yunjin Chen}
received the B.Sc. degree in applied physics from Nanjing University of Aeronautics and Astronautics, Jiangsu, China, and 
the M.Sc. degree in optical engineering from National University of Defense Technology, Hunan, China, in 2007 and 2009, 
respectively. Since September 2011, he has been pursing the Ph.D degree at the 
Institute for Computer Graphics and Vision, Graz University of Technology. His current research interests are 
learning image prior model for low-level computer vision problems and convex optimization. 
\end{IEEEbiography}
\vspace*{-2cm}
\begin{IEEEbiography}[{\includegraphics[width=1in,height=1.25in,clip,keepaspectratio]
{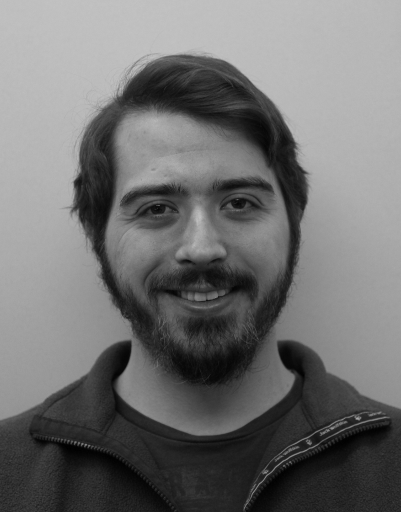}}]{Ren{\'e} Ranftl}
received the B.Sc. degree and M.Sc. degree in
Telematics from Graz University of Technology in 2009 and 2010,
respectively. Since 2010 he has been pursuing the Ph.D. degree 
at the Institute for Computer Graphics and Vision, Graz University
of Technology. His current research interests include image sequence 
analysis, 3D reconstruction and machine learning.
\end{IEEEbiography}
\vspace*{-2cm}
\begin{IEEEbiography}[{\includegraphics[width=1in,height=1.25in,clip,keepaspectratio]
{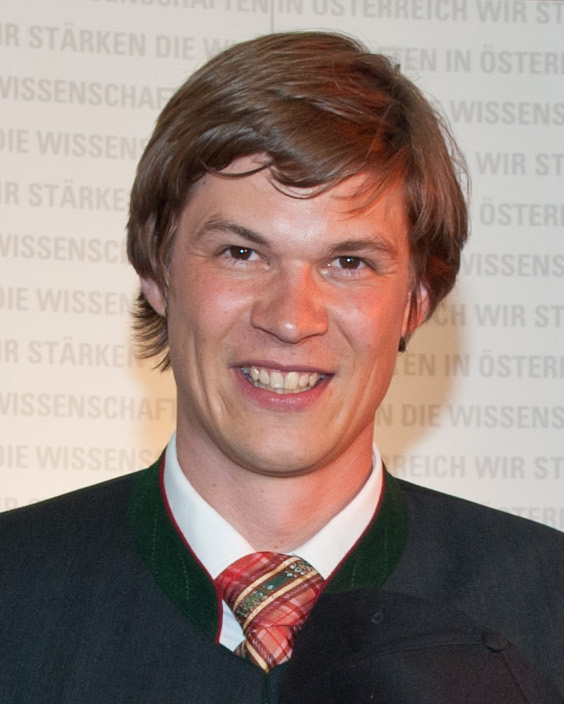}}]{Thomas Pock}
received a M.Sc and a Ph.D degree in Telematik from Graz University of Technology in 2004 and 2008, respectively. He is currently employed as an Assistant Professor at the Institute for Computer Graphics and Vision at Graz University of Technology. In 2013 he received the START price of the Austrian Science Fund (FWF) and the German Pattern recognition award of the German association for pattern recognition (DAGM). His research interests include convex optimization and in particular variational methods with application to segmentation, optical flow, stereo, registration as well as its efficient implementation on parallel hardware. 
\end{IEEEbiography}

\end{document}